\documentclass[fleqn,10pt]{SelfArx} 

\usepackage{lipsum} 
\usepackage{hyphenat}
\usepackage{xurl}

\usepackage{booktabs} 
\usepackage{appendix}
\usepackage{siunitx} 
\usepackage{xparse} 
\usepackage{subcaption}
\usepackage[utf8]{inputenc}
\usepackage[T1]{fontenc}
\definecolor{color1}{RGB}{0,0,90} 
\definecolor{color2}{RGB}{0,20,20} 

\usepackage{hyperref} 
\hypersetup{hidelinks,colorlinks,breaklinks=true,urlcolor=color2,citecolor=color1,linkcolor=color1,bookmarksopen=false,pdftitle={Title},pdfauthor={Author}}

\JournalInfo{Draft Version} 
\Archive{.} 

\PaperTitle{Shaping History: Advanced Machine Learning Techniques for the Analysis and Dating of Cuneiform Tablets over Three Millennia} 

\Authors{Danielle Kapon\textsuperscript{1}*, Michael Fire\textsuperscript{2}, Shai Gordin\textsuperscript{3}\textsuperscript{4}} 
\affiliation{\textsuperscript{2}\textit{Department of Software and Information Systems Engineering, Ben-Gurion University of the Negev, Be'er Sheva, Israel}} 
\affiliation{\textsuperscript{2}\textit{Department of Software and Information Systems Engineering, Ben-Gurion University of the Negev, Be'er Sheva, Israel}} 
\affiliation{\textsuperscript{3}\textit{Department of Land of Israel Studies and Archaeology, Digital Pasts Lab, Ariel University, Ariel, Israel}} 
\affiliation{\textsuperscript{4}\textit{Digital Humanities and Social Sciences Hub, Open University of Israel, Ra'anana, Israel}} 
\affiliation{*\textbf{Corresponding author}: kapond@post.bgu.ac.il} 

\newcommand{\keywordname}{Keywords} 

\Abstract{
Cuneiform tablets, emerging in ancient Mesopotamia around the late fourth millennium BCE, represent one of humanity's earliest writing systems. Historically characterized by wedge-shaped marks on clay tablets, these artifacts provided insight into Mesopotamian civilization across administrative, commercial, legal, literary, and scientific domains. The traditional analysis and dating of these tablets still mainly rely on manual, subjective appreciation of shape and writing style. This approach, however, leads to many uncertainties in pinpointing the exact period the tablets originated from. Recent advances in digitization have revolutionized the study of cuneiform by enhancing accessibility and analytical capabilities. Our research uniquely focuses on the physical shapes of tablets as significant indicators of their historical periods, diverging from most studies that concentrate on textual content. Utilizing an unprecedented dataset of over 94,000 images from the Cuneiform Digital Library Initiative (CDLI) collection, our effort applies deep learning methods to the classification of cuneiform tablets, covering more than 3,000 years of history. By leveraging statistical, computational techniques, and innovative generative modeling through Variational Auto-Encoders (VAEs), we achieve substantial advancements in the automatic classification of these ancient documents, by focusing on the tablets' silhouette as a main predictor. Our classification approach begins with a Decision Tree using the height-to-width ratios of the tablets as the sole predictor which achieved 8\% macro F1-score, and culminates with a ResNet50 model, which achieved a 61\% macro-F1 score for tablet silhouettes. Additionally, we introduce a novel tool-set, powered by VAEs, to enhance the interpretability of our models and enable researchers to explore the changes in tablet shapes across different eras and genres. This research contributes to the fields of document analysis and diplomatics by demonstrating the value of large-scale data analysis combined with statistical methods, also within specific historical periods. These telescopic and microscopic insights offer valuable tools for historians and epigraphists, enriching our understanding of both cuneiform tablets themselves and the cultures that produced them.
}
\begin{document}

\flushbottom 
\maketitle 
\thispagestyle{empty} 
\vfill
\section{Introduction} 

\addcontentsline{toc}{section}{Introduction} 
The emergence of writing, specifically the cuneiform logo-syllabic writing system impressed with a reed stylus on clay tablets, finds its roots in ancient Mesopotamia, one of the few places where this occurred independently \cite{woods2010visible, coulmas2008typology, brokaw2022homo, daniels1996study, Cammarosano_2014}. The Sumerians, followed by the Elamites, Akkadians, Hurrians, Hittites, and many other ancient Near Eastern cultures, used this medium to produce a vast collection of writings, including archives, literature, and scientific texts, many of which survive due to the durability of clay \cite{walker1987cuneiform}. While much writing has been lost, those on durable materials like clay, metal, and stone have withstood the harsh Middle Eastern climate \cite{walker1990reading}. Despite their durability, clay tablets can deteriorate if not properly conserved, making restoration challenging \cite{fetaya2020restoration}. Spanning from the fourth to the first millennium BCE and written in at least a dozen languages, these tablets offer crucial insights into ancient societies across regions from Iran to Egypt, the Levant, and Anatolia \cite{hagelskjaer2022deep, altorientalistik2010umfang, RattenborgSmidtJohanssonMelinKronsellNett+2023+178+205}. These cuneiform inscriptions, encompassing various genres, serve as a window into ancient societies, their politics, history, law, and sciences \cite{woods2010visible}. However, processing hundreds of thousands of tablets remains a challenge due to the complexity of the content of these tablets, and the relatively small number of experts involved \cite{watkins2003digital, kumar2003digital, cohen2004iclay}. 

\begin{figure*}
    \centering
    \begin{subfigure}[T]{0.28\textwidth}
        \centering
        \includegraphics[width=\textwidth]{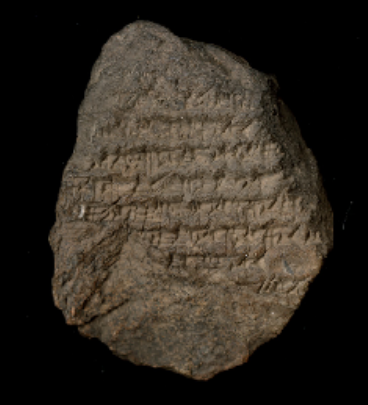}
        \caption{Neo-Assyrian\\ (ca. 911-612 BC), \\divination \cite{2023K}}
        \label{fig:tableta}
    \end{subfigure}
    \begin{subfigure}[T]{0.14\textwidth}
        \centering
        \includegraphics[width=\textwidth]{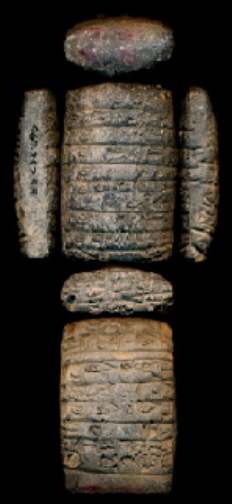}
        \caption{Ur III \\ (ca. 2100-2000 BC), \\ administrative \cite{2002Ontario}}
        \label{fig:tabletb}
    \end{subfigure}
    \begin{subfigure}[T]{0.235\textwidth}
        \centering
        \includegraphics[width=\textwidth]{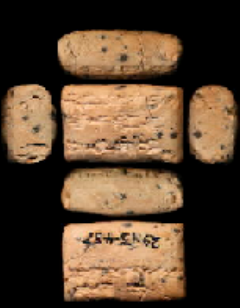}
        \caption{Middle Babylonian\\ (ca. 1400-1100 BC), \\administrative \cite{2023UM}}
        \label{fig:tabletc}
    \end{subfigure}
    \begin{subfigure}[T]{0.265\textwidth}
        \centering
        \includegraphics[width=\textwidth]{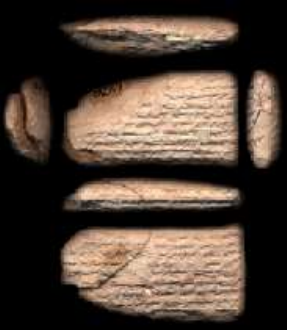}
        \caption{Neo-Babylonian\\ (ca. 626-539 BC),\\ scholarly or scientific \cite{2023CBS}}
        \label{fig:tabletd}
    \end{subfigure}
    
    \caption{A sample of some of the variety of shapes and sizes of cuneiform tablets texts: (a) Neo-Assyrian; (b) Ur III; (c) Middle Babylonian; (d) Neo-Babylonian.}
    \label{fig:tablets}
\end{figure*}

Dating cuneiform tablets presents ongoing challenges. While features like clay composition, size, shape, stylistic elements, sign forms, text type and content, associated finds, and museum "archaeology" offer solid clues, not many standardized methods based on measurable metrics in a large-scale statistical distribution have been established for suggesting their chronological sequence~\cite{biggs1973regional, Fales_2003, kloekhorst2019hittite, Buchanan_2023}.\footnote{Another method that produces a measurable metric based on a large statistical distribution is archaeomagnetic dating, which depends on whether the clay artifact was baked in antiquity. This has been demonstrated in the analysis of 32 inscribed baked bricks from Mesopotamia (3rd–1st millennia BCE), providing high-resolution geomagnetic intensity data. Whereas our method is purely shape-based and has disadvantages in the case of fragmentary preservation, some archives, cities, and areas have consistent fragmentation states that can suggest possible dating~\cite{howland2023exploring}} Significant discrepancies among scholars may occur despite handling tablets from eras that have a vast collection of dated documents~\cite{biggs1973regional, kloekhorst2019hittite, matskevich20181}; Literary compositions, for example, some of which are extent in many copies and in different manuscript traditions, have had their manuscripts variously dated to a period covering almost a thousand years~\cite{millard1969atra, van2022literarische, marquez2016two}. This underscores the challenges in comprehending the evolution of the oldest recorded written language \cite{van2009century}.

The information age, however, has ushered in a transformative approach to this challenge. Through advanced 2D and 3D photographic techniques, the digitization of cuneiform tablets has led to the creation of public databases~\cite{CDLI2023Home, opendanes2024, data/IE8CCN_2019}. This digitization paves the way for the application of cutting-edge computer vision and machine learning techniques, revolutionizing how scholars and researchers approach the study of these ancient texts.

Prior research in the application of computational methods for dating cuneiform tablets has primarily focused on classifying their historical period based on the physical shape of the signs on the tablets~\cite{9257733, YugayPa}, performing Optical Character Recognition (OCR) on their text \cite{bogacz2022digital}, or analyzing their content and genre \cite{pmlr-v88-kriege18a, gordin2020reading, DBLP:journals/corr/abs-2009-10794, homburg-chiarcos-2016-word, dencker2020deep, mahmood2023classifying}. However, these studies have left a critical gap unaddressed: the global interpretability of the models used. It is crucial to inquire into the domain-specific reasoning behind the models' decisions, a task typically carried out by historians or epigraphists. Furthermore, to our knowledge, there has not yet been a study of tablet shapes that tackles a dataset as expansive across time and space as the one we propose to analyze.

Our study introduces a novel approach by employing machine learning to automatically date cuneiform documents, focusing on the physical shape of tablets as a primary indicator of their historical period. This method aims to reduce subjectivity, offering a measurable and interpretable framework for determining document age, thereby enhancing the precision of dating ancient texts. The center of our research revolves around understanding the shape-related features that make each historical period different than the next. We are also interested in being able to show what the typical cuneiform tablet from a certain period, or of a certain genre, looks like, and offer digital analysis tools.

\begin{figure*}[ht]
     \centering
     \begin{subfigure}[b]{0.54\textwidth}
         \centering
         \includegraphics[width=\textwidth]{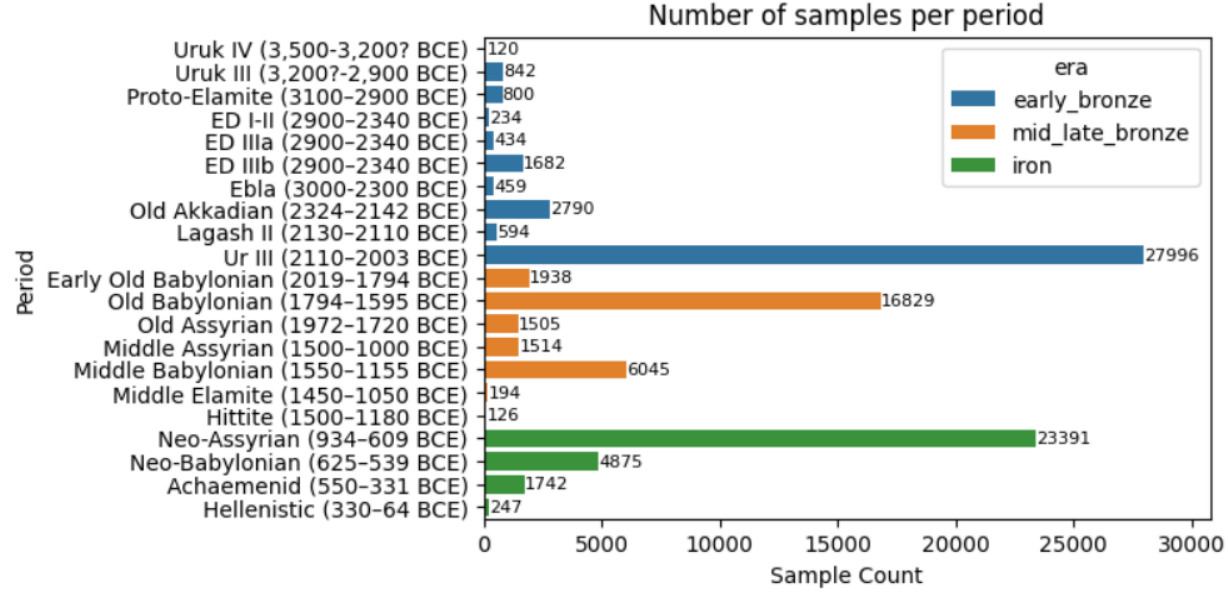}
         \captionsetup{justification=centering}
        \caption{Number of Samples per Historical Period}
         \label{fig:nsamples_per_period}
     \end{subfigure}
     \hfill
     \begin{subfigure}[b]{0.45\textwidth}
         \centering
         \includegraphics[width=\textwidth]{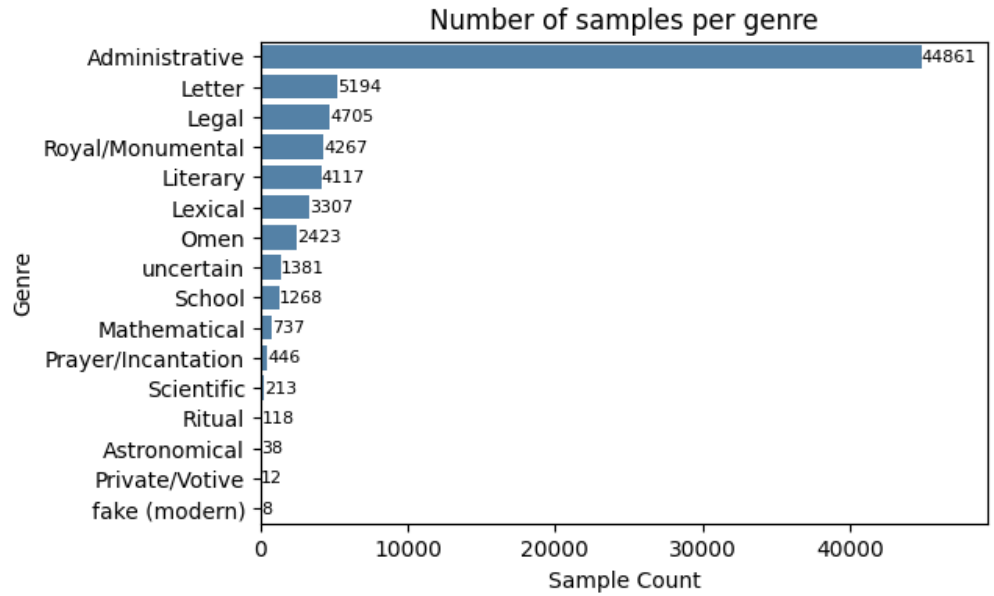}
         \captionsetup{justification=centering}
         \caption{Number of Samples per Genre}
         \label{fig:nsamples_per_genre}
     \end{subfigure}
        \caption{Number of Samples per Period and Genre}
        \label{fig:nsamples_per_period_and_genre}
\end{figure*}

Initially, our focus was on preprocessing 94,936 tablet images from the CDLI catalogue \cite{CDLI2023Home} (period and genre distribution can be seen in Figure \ref{fig:nsamples_per_period_and_genre}). We applied binary masks to grayscale images to generate black-and-white versions, effectively isolating shape-related information by eliminating extraneous visual elements. Following this preprocessing step, we performed an exploratory data analysis (EDA) to scrutinize the 2D geometric properties of the masked tablets, paying particular attention to their height-width ratios. This phase aimed to enhance our understanding of the tablets' shapes in preparation for subsequent classification tasks.

Our approach for analyzing cuneiform tablets involved multiple stages, beginning with a tablet height-width ratio analysis to measure tablet types over time and look into prominent statistical trends, and is the first large scale collection of such measurements, providing a robust basis for further analysis. We proceeded with tablet period classification through various models, including the use of DINOv2~\cite{oquab2023dinov2} and a basic Convolutional Neural Network (CNN)~\cite{6313076}, which yielded a macro F1-score to 49\% for masked images and 61\% for grayscale images. A significant leap in accuracy was achieved by fine-tuning the pre-trained ResNet50 model, tailored for shape-based classification, achieving a macro-F1-score of 71\% on grayscale images and 61\% on the masked tablets. Notably, the performance on masked images demonstrated that approximately 86\% of the essential classification information was preserved, affirming the relevance of our focus on shape-related features.

We enhanced our analysis of cuneiform tablet shapes across different historical periods using Variational Auto-Encoders (VAEs)~\cite{kingma2013auto}, chosen for their ease of training, sample diversity, and mainly - their interpretability. These generative models allowed us to examine and categorize the unique shape characteristics of the tablets by historical period and genre. By analyzing the encoded representations from the VAE's bottleneck layer vectors, we directly observed the relationships between tablet shapes and their historical contexts. This approach not only demonstrated the utility of VAEs in understanding the evolution of tablet shapes across different periods and genres but also established a robust framework for classifying and comprehending cuneiform tablets, showing how their shapes have progressed from one period to another, among various other applications.

\begin{figure*}[ht]
\centering
\begin{subfigure}[T]{0.56\textwidth}
    \centering
    \includegraphics[width=\textwidth]{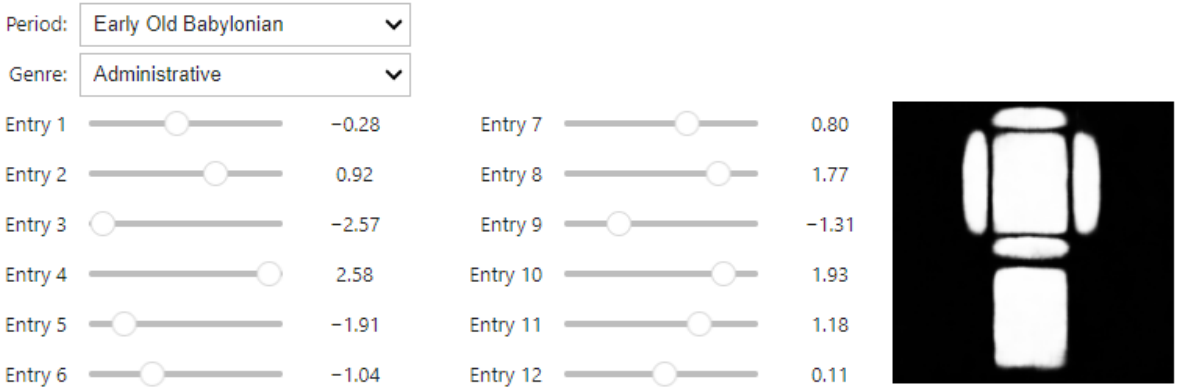}
    \caption{A widget that provides an interactive platform for users to examine a sample tablet from selected genres and historical periods. It facilitates an exploration of the various factors encoded in the VAE's bottleneck layer, shedding light on the attributes that define the stylistic and temporal aspects of the tablet, and enhancing understanding of its historical and cultural context.}
    \label{fig:widget_bottleneck}
\end{subfigure}
\hfill
\begin{subfigure}[T]{0.40\textwidth}
    \centering
    \includegraphics[width=\textwidth]{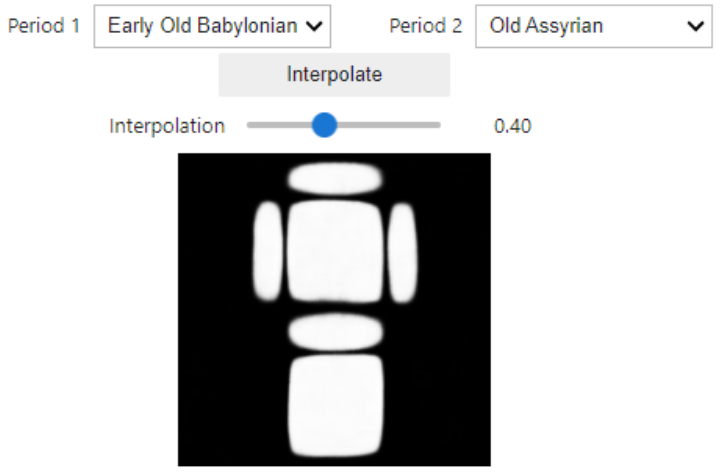}
    \caption{A widget that enables users to interpolate between the mean tablet representations of two distinct periods, as captured in the bottleneck layer of a VAE. By adjusting the slider, users can visually explore the evolution of tablet designs across periods, highlighting the average distinguishing features of tablet shapes between the chosen historical periods. The figure shows a tablet consisting of 40\% Old Assyrian and 60\% Early Old Babylonian tablet}
    \label{fig:interpulation_widget}
\end{subfigure}
\caption{Interactive widgets for exploring tablet characteristics and evolution.}
\label{fig:widgets_for_tablet_exploration}
\end{figure*}

We were able to show that the shape of the tablets, even when represented in 2D, holds a large portion of the information regarding cuneiform tablets dating to historical periods. We were able to use the bottleneck layer of a VAE to show patterns in tablet shapes and provide visual tools to describe the average characteristics of tablets from different eras, periods, and genres, cluster tablets, and explore similarities. (Figure
\ref{fig:widgets_for_tablet_exploration}).

The key contributions presented in this paper are as follows:
\begin{itemize}
    \item We conducted a thorough shape analysis and classification over the largest dataset of cuneiform tablets to date, spanning more than 3,000 years.
    \item We demonstrated that 2D silhouette representations of tablet shapes hold critical information for associating cuneiform tablets with their specific historical periods and leveraged these findings into period classification.
    \item We leveraged VAEs to uncover patterns within tablet shapes, aiding in distinguishing features and facilitating the development of practical visual aids for analysis.
    \item We analyzed the latent features extracted by VAEs, and introduced methods for characterizing average features across eras, periods, and genres, enabling automated clustering and detailed comparison of tablets.
    \item We developed two widgets to enhance cuneiform tablet analysis: one for interactive shape analysis and dating and another for visualizing design evolution over time, utilizing deep learning and generative modeling to offer new insights into these ancient artifacts.
\end{itemize}

Together, these contributions provide tools for historians and epigraphists to accelerate and refine their research on tablet dating, as well as to gain global insights regarding the relationship between tablet shape and various genres, periods, and places of origin.

The remainder of this paper is organized as follows: Section \ref{related_work} presents existing research on tablet-based dating, interpretability of machine learning models, and the application of generative models for interpreting data. Next, Section \ref{methods} details our research objectives and methodology. We begin by outlining our core approach to tablet classification. We then introduce a novel use of generative deep neural networks, specifically VAEs, to enhance interpretability through analysis of the latent space. Section \ref{results} presents the study's findings, and Section \ref{discussion}, highlights key insights into tablet dating and the interpretability through VAEs. The paper concludes in Section \ref{conclusions}, summarizing our contributions and discussing future research directions.


\section{Related Work} \label{related_work}

\subsection{Tablet Dating}
Due to the absence of contemporary palaeography manuals and the scarcity of exact methods to measure signs, crucial data cannot be retrieved for the numerous cuneiform records that lack clear archaeological context. 
The shape of a cuneiform tablet, which is part of the systematic study of external document features called diplomatics, is a significant feature that assists in estimating the historical period of its creation, and can also be governed by scholarly, social, and political factors \cite{kloekhorst2019hittite,Veenhof1986archives, Charpin_2002,Taylor2011tablets}. Different tablet shapes (also referred to as format) may be attributed to the period of writing, but also to the context and their genre \cite{waal2016hittite, hackl2021artaxerxes, matthews2013writing}. For example, Podany \cite{podany1991middle} and Walker \cite{walker2021artaxerxes} found that tablets containing contracts had different height-width ratios according to their date. Hackl \cite{hackl2021artaxerxes} identified that tablet shapes in the Late Babylonian Tattannu archive dramatically change when arranged by groups according to their functionality and over time (late 6\textsuperscript{th} - early 4\textsuperscript{th} centuries BCE). Moreover, certain document types have a cushion-like shape, a generally more circular form, curved corners or sides, or a generally irregular shape \cite{lecompte2016observations}. These shapes are attributes that can be seen even in texts that are partially preserved. The significance of shape analysis in other artifacts, such as pottery, underscores the importance of such features in comprehensively understanding the origins of artifacts and measuring their style, both within the same historical setting (synchronically), over time (diachronically), and across space (geographically) \cite{parisotto2022unsupervised, CARDARELLI2022105640}. This recognition highlights the potential for applying similar methodologies to analyze and classify cuneiform tablets.

\subsection{Using Machine Learning for Tablet Dating}

As far as we are aware, only one study was conducted to assist scholars in the task of tablet dating based on cuneiform tablet images, incorporating machine learning and mainly deep learning techniques to classify the period in which a tablet was written or the text in the tablets. Bogacz and Mara \cite{bogacz2020period} proposed using a neural network for prediction, using the mesh 3D representation of the tablet, over four time periods - ED IIIb, Ur III, Old Babylonian, and Old Assyrian.

An example for a similar study in ancient Greek, is Assael and Sommerschield et al. \cite{assael2022restoring}, which used an input of character embeddings, word embeddings, and positional embeddings, as part of a transformers-based architecture for a restoration task, which can assist with the classification task of dating. For similar studies in other ancient languages see Sommerschield et al. \cite{sommerschield2023machine}.

These studies demonstrated a commendable capacity to categorize the tablets based on their respective periods; However, they used a small dataset containing a small number of periods, particularly periods with a large number of processed texts. Furthermore, the factors that drove the models' classification process are uncertain, and it is unclear which features were responsible for assigning them to a specific period. Also, many assumptions regarding dates recorded in existing datasets may be biased or based on circular reasoning, which would in turn create issues when used for training machine learning models.

Our study provides a significant contribution to this discussion, by leveraging a comprehensive collection of tablets from various periods spanning several millennia. Even though the dataset we utilize, i.e., the 2D shapes of tablets sourced from the CDLI database, presents several challenges, such as the uneven distribution of data points, as well as possible errors in recorded dates, it is first and foremost used to create a large scale statistical baseline of tablet shapes. Then, machine learning methods are used to extract analytical features in tablet shapes, which could be used to identify stylistic connections and potential features for tablet dating.

\subsection{Interpretability of Machine Learning and Deep Learning models}

In many studies, explainability techniques are required to comprehend the decision-making process of Machine Learning (ML) and Deep Learning (DL) algorithms \cite{spinner2020explainer}. As models become more complex, it becomes increasingly challenging to elucidate their decisions \cite{duran2021afraid}. Explainable AI (XAI) is a field of AI solutions that seeks to reveal the internal workings of ML/DL models to users but is also a big part of the validation process before deploying a model to production \cite{arrieta2020explainable, schockaert2020vae, murdoch2019definitions}. The definitions of interpretability and explainability in AI differ significantly as the field is still in its early stages, as observed in various works reviewed by Salahuddin et al. \cite{salahuddin2022transparency}.\\
Local Interpretable Model-agnostic Explanations (LIME) is one solution for explaining complex models. LIME employs a locally accurate multi-regression model to approximate the behavior of a model, working with both tabular and imagery data \cite{ribeiro2016should, magesh2020explainable}. Shapley Additive Explanations (SHAP) is another popular solution \cite{lundberg2017unified}. It employs game theory to assess the significance of each feature in the dataset and its contribution to a coalition. Several model-agnostic frameworks have been developed that rely on the perturbation technique. This technique involves assessing how much a feature contributes to the model's prediction by observing how changes in the feature's value affect the prediction outcome. This approach is detailed in works by Fong et al. and Vu et al. \cite{fong2019explanations, fong2017interpretable, vu2019evaluating}, where they explore how altering feature values impacts model predictions to understand the importance and influence of each feature. Despite being utilized for various applications and being revolutionary in the field of XAI, the techniques above provide only a local and faithful explanation for each sample and are unable to offer a comprehensive understanding of the decisions made by a black-box model \cite{treppner2022interpretable}. \\
Our research aims to comprehend the overall characteristics contributing to the classification of cuneiform document images, which requires using alternative techniques.

\subsubsection{Analysis by Synthesis}
The emergence of generative models has significantly impacted both the academic community and industry applications in recent times, specifically in the field of computational Archaeology \cite{parisotto2022unsupervised, CARDARELLI2022105640, bickler2021machine}. Utilizing these models, particularly for image synthesis through CNNs, enhances our ability to reveal the underlying semantic structures of the network, offering a deeper insight into the defining shape characteristics that contribute to the uniqueness of tablet production across different periods \cite{salahuddin2022transparency, treppner2022interpretable}. To provide interpretations, scholars aim to utilize the latent space of CNNs, which denotes the condensed form of the input image fed into the model. The process of disentangling representations in the latent space aims to identify the most impactful characteristics in the data, with the expectation that these traits will be uncorrelated with one another \cite{higgins2018towards}.

The last decade has seen significant advancements in generative models, with recent progress unlocking insights into their inner workings for enhanced interpretability.
One such type of models is Generative Adversarial Networks (GANs), published in 2014, which are a class of neural network architectures composed of two distinct models: a generator and a discriminator. These models engage in a game-like scenario where the generator aims to create data that is indistinguishable from real data, while the discriminator evaluates whether the given data is real or produced by the generator. This dynamic allows GANs to generate highly realistic data samples \cite{goodfellow2014explaining}. 
Recent advancements, particularly in variations of GANs such as StyleGAN, have showcased the ability to disentangle and manipulate the latent space. This development allows for greater control over the characteristics of the generated images, enhancing the capacity to generate data with specific attributes or styles \cite{karras2019style, chen2022s, wu2021stylespace, schutte2021using}.

VAEs refine the autoencoder concept by introducing a mechanism to better navigate the latent space, enhancing the diversity and relevance of generated images. Unlike standard autoencoders, which may struggle with diversity due to their reconstruction loss function, VAEs incorporate a Kullback-Leibler (KL) divergence loss. This additional loss function measures the deviation from a prior distribution, ensuring a more structured and continuous latent space. It allows VAEs to not just replicate input images but to also generate new, domain-appropriate samples by adding parameters $\mu$ and $\sigma$, creating a probabilistic space from which to draw. This approach significantly mitigates the issue of variability and enables the traversal between different image classes more effectively ~\cite{kingma2013auto, salahuddin2022transparency,treppner2022interpretable,
spinner2018towards, blaschke2018application}.

In their initial introduction, VAEs outperformed existing models, yet further analysis revealed a significant limitation: the entanglement of their latent variables. This entanglement, resulting from complex interconnections within the neural network layers, means that the latent space—where the model encodes information—is not clearly organized. Individual variables are not easily distinguishable, complicating efforts to understand or modify the effects of specific features on the model's outputs~\cite{mathieu2019disentangling, ainsworth2018oi}. 

This is especially important for us, as in our research we wish to be able to give clear explanations regarding the tablet shape features leading to the period classification. Over the years, several disentanglement techniques were introduced to tackle this issue. Schockaert et al.~\cite{schockaert2020vae} created VAE-LIME, a local interpretability technique based on the ideas of LIME, but incorporating VAEs as a part of the perturbation process. VAE-LIME showed an improved fidelity of the local model created to interpret black-box models. Another model called pi-VAE \cite{zhou2020learning}, attempted to combine latent-based and regression-based approaches for interpreting neural population data; oi-VAE \cite{ainsworth2018oi} focused on non-linear explanations, offering higher expressive power, and introducing sparsity-inducing penalty, assisting the disentanglement or the latent representations; $\beta$-VAE \cite{mathieu2019disentangling, chen2018isolating, burgess2018understanding}, suggested an augmentation factor $\beta$ for the KL-divergence loss function portion, and was able to uncover multiple factors of variation in the data and produce representations that are more comprehensive and cleanly disentangled, while outperforming alternative models. Other works \cite{mathieu2019disentangling}, analyzed the characteristics of disentanglement of $\beta$-VAE and offered improvements to the mentioned techniques to create a more flexible and generic framework. NVAE, introduced by Vahdat et al.~\cite{vahdat2020nvae}, showed advanced VAE models with a hierarchical, multi-scale architecture, enhancing image quality and training stability.

Denoising Diffusion Probabilistic Models (DDPMs), published in 2021, are an evolving frontier in generative models, with enhancements broadening their scope from image generation to interpretability, though the latter is less examined ~\cite{ho2020denoising,ho2022cascaded,sohl2015deep, nichol2021improved,nichol2021glide,song2020denoising,tseng2022hierarchically}. They employ Markov chains for the incremental introduction of noise into the data and leverage U-Nets for the denoising phase, systematically removing noise to regenerate images that resemble the original training data. Other approaches for explainability used Auto Encoders (AEs), and specifically VAEs for understanding the connection between inputs and their underlying contextual features \cite{treppner2022interpretable, biffi2020explainable, clough2019global, utkin2021combining, uzunova2019interpretable}, as well as manipulating the outputted images based on the latent disentangled attributes.

\subsubsection{The Generative Learning Trilemma}
When we come to choose a generative model to assist with our main task of analysis of cuneiform tablet shape features that contribute to its chronological origin, we need to consider a few issues.

While GANs tend to produce photo-realistic images as output, they tend to be hard to train due to the instability of their loss function, which is composed of a generator loss and a discriminator loss, competing with each other \cite{salahuddin2022transparency, treppner2022interpretable, goodfellow2020generative}. It also lacks variability in its generated images, as it is improbable to discover a latent value that the decoder can utilize to create valid outputs.
Despite occasionally surpassing GANs in generating high-quality samples \cite{dhariwal2021diffusion, rombach2022high}, DDPMs encounter significant hurdles due to their prolonged sampling times and challenging training processes, requiring thousands of network evaluations \cite{ho2022cascaded, nichol2021glide, xiao2021tackling}. This complexity makes them less suitable for some real-world applications and reduces their potential for interpretation.
VAEs are also much easier to train than GANs, have higher stability, and are more efficient; However, they lack the photo-realistic characteristics of AE, GANs, and Diffusion models \cite{spinner2018towards, xiao2021tackling}, due to the penalty meant to restrict the network from learning latent representations that deviate significantly from a standard normal distribution (measured by the addition of the KL-divergence loss to the loss function). 

This issue is addressed as the "Generative Learning Trilemma" \cite{xiao2021tackling}; This is because generative models struggle to excel simultaneously in three important requirements: easy and fast training process, sampling diversity, and high sampling quality. As of now, no model was able to perfect all three of those key requirements. While producing high-quality images with high sampling diversity, Diffusion models tend to be hard and expensive to train \cite{dhariwal2021diffusion, rombach2022high}, hence are inapplicable for many real-life problems.

\section{Methods} \label{methods}
Our research explores the classification of cuneiform tablets' chronological periods through their shapes. Initially, we apply classification models to assess the predictive power of tablet silhouettes for dating purposes. This foundational analysis paves the way for our main objective: employing generative models to synthesize tablet shapes, thus advancing period classification by merging traditional methods with innovative analysis-by-synthesis techniques. This synthesis-driven methodology allows us to explore and identify defining characteristics of tablets across different historical periods more deeply. In this section, we describe the data sources, processing steps, methodologies, and evaluation metrics included in our research. We highlight our integrated approach that combines direct classification with analysis by synthesis to uncover new insights into the chronology of cuneiform records.

\subsection{Data Acquisition}

Our work utilizes a comprehensive dataset of cuneiform table\-ts sourced from the CDLI
API~\cite{CDLI2023About}. This vast collection, spanning various museums and collections and representing the history of writing (3350 BC - pre-Christian era), encompasses over 360,000 digitized catalog entries. For our specific use, we selected 94,936 artifacts equipped with both an image and a historical period classification that was provided in the dataset by domain experts \cite{CDLI2024Periods}. Each image has additional information like genre and origin, enriching our understanding of its characteristics. A sample of these tablets can be seen in Figure~\ref{fig:tablets}. 

The dataset encompasses a diverse range of periods, categorized into three major eras, as can be seen in Figure~\ref{fig:nsamples_per_period}:\footnote{For the sake of convenience all periodization is estimated based on the Middle Chronology \cite{Pruzsinszky2009, sallaberger2015arcane}.}

\begin{enumerate}
  \item \textbf{Early Bronze Age (3rd Millennium BCE)} \cite{sallaberger2015arcane}: Uruk IV (3,500-3,200? BCE), Uruk III (3,200?-2,900 BCE), Proto-Elamite (3,100-2,900 BCE), ED I-II (2,900-2,340 BCE), ED IIIa (2,900-2,340 BCE), ED IIIb (2,900-2,340 BCE), Ebla (3,000-2,300 BCE), Old Akkadian (2,324-2,141 BCE), Lagash II (2,130-2,110 BCE), and Ur III (2,110-2,003 BCE).
  \item \textbf{Middle-late Bronze Age (2nd Millennium BCE)} \cite{hoflmayer2022establishing, reculeau2022assyria}: Early Old Babylonian (2,019–1,794 BCE), Old Babylonian (1,794-1,595 BCE), Old Assyrian (1,972-1,720 BCE), Middle Assyrian (1,500-1,000 BCE), Middle Babylonian (1,550-1,155 BCE), Middle Elamite (1,450-1,050 BCE), and Hittite (1,500-1,180 BCE).
  \item \textbf{Iron Age (1st Millennium BCE):} Neo-Assyrian (934-509 BCE), Neo-Babylonian (625-539 BCE), Achaemenid (550-331 BCE), and Hellenistic (330-64 BCE).
\end{enumerate}
                             
These period classifications aided our supervised image classification algorithm and the validation of our generation process. The images served as the core data for our analysis, classification, and synthesis tasks.

\subsection{Data Preprocessing}
During the data acquisition, the scraped images are stored in a grayscale format and resized proportionally to 512x512 pixels (adding a black background to rectangular images) to reduce model complexity. We performed this image preprocessing because we do not require the intricate details of the tablet, allowing us to use images of lower resolution. 

\begin{figure}[htbp]
\centering 
\begin{subfigure}[T]{.22\textwidth}
  \centering
  \includegraphics[width=\linewidth]{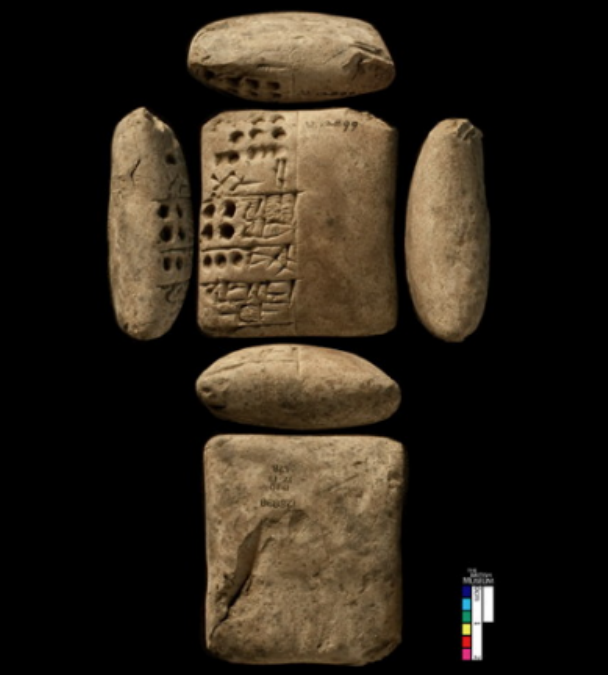}
  \caption{A tablet featuring a color and scale bar. ED I-II (ca. 2900-2700 BC), Administrative genre.}
  \label{fig:color-bar-tablet}
\end{subfigure}
\begin{subfigure}[T]{.22\textwidth}
  \centering
  \includegraphics[width=\linewidth]{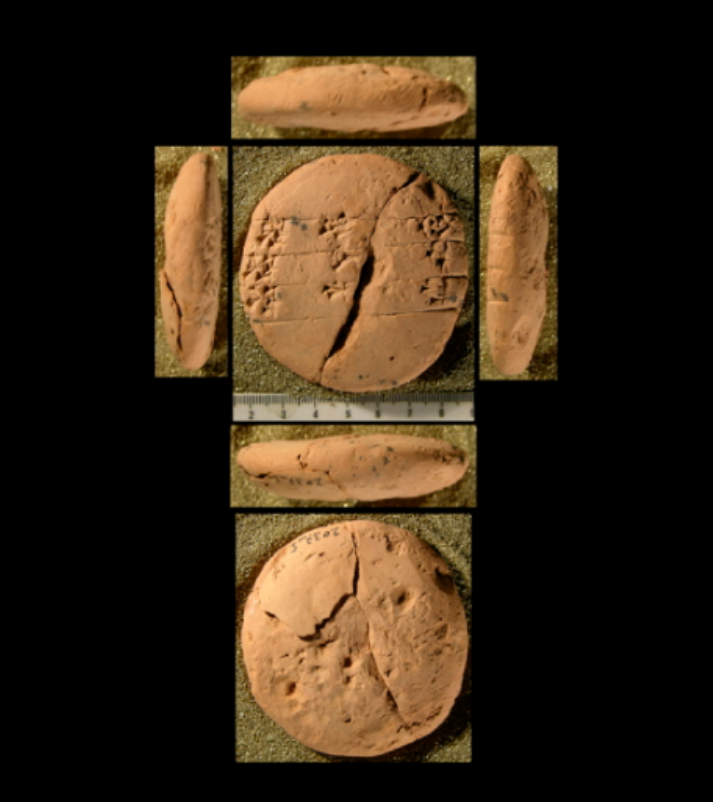}
  \caption{A tablet originally photographed on a non-black background. Early Old Babylonian (ca. 2000-1900 BC), Lexical genre.}
  \label{fig:non-black-back-tablet}
\end{subfigure}
\caption{Tablets addressed in the preprocessing stage of the research}
\label{fig:tablets-sidebyside}
\end{figure}

To effectively isolate the shape of ancient tablets from their images for analysis or digital archiving, our methodology involves creating binary masks that emphasize the tablet's silhouette. Initially, we remove irrelevant features such as scale bars and logos, employing both manual and automated techniques to ensure the images are free of distractions. This process is critical as it ensures we will not get a bias due to these external elements and allows focus on the tablet, as illustrated in Figure~\ref{fig:tablets-sidebyside} \cite{2023UET}. Additionally, we exclude images that do not provide a clear view of the tablet, particularly those taken against non-black backgrounds, as shown in Figure \ref{fig:non-black-back-tablet}~\cite{2023CUSAS}, to guarantee the integrity of our dataset.

\begin{figure}[htbp]
    \centering
    \begin{subfigure}[b]{0.32\linewidth}
        \centering
        \includegraphics[width=\textwidth]{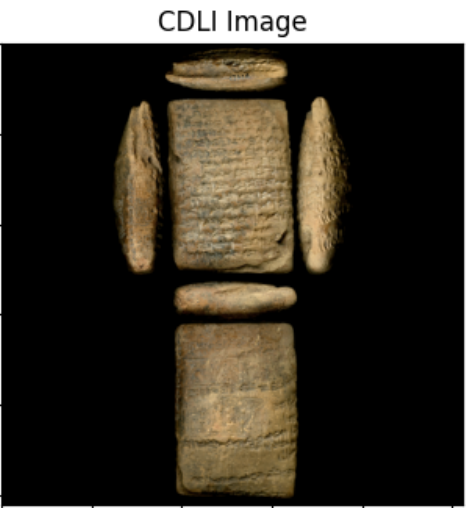}
        \caption{Original image from CDLI~\cite{2023RA}.}
        \label{fig:transformation_a}
    \end{subfigure}
    \hfill
    \begin{subfigure}[b]{0.32\linewidth}
        \centering
        \includegraphics[width=\textwidth]{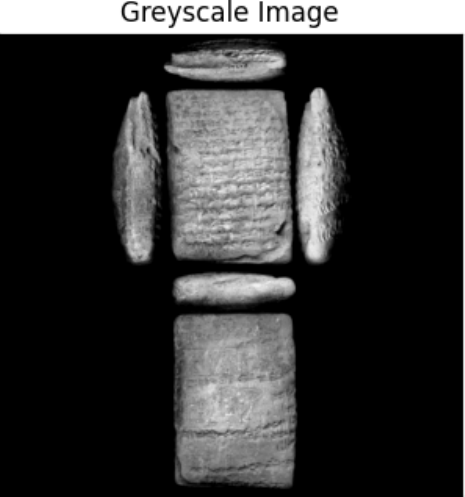}
        \caption{Grayscale image after preprocessing.}
        \label{fig:transformation_b}
    \end{subfigure}
     \hfill
    \begin{subfigure}[b]{0.32\linewidth}
        \centering
        \includegraphics[width=\textwidth]{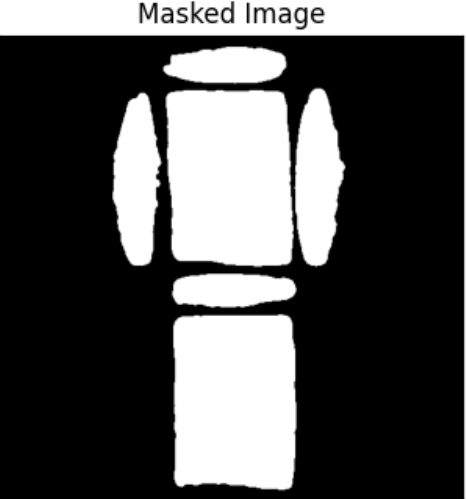}
        \caption{Masked image after preprocessing.}
        \label{fig:transformation_c}
    \end{subfigure}

    \caption{Illustration of the process for extracting the largest component of a tablet.}
    \label{fig:mask_preprocess}
\end{figure}

We apply Gaussian blur to the grayscale images to obtain a binary image representing the tablet shapes, followed by carefully selecting a binary threshold.\footnote{The Gaussian blur and binarization thresholds were manually selected and tested on several hundred random images. This process aimed to prevent the darker areas of the tablets from turning completely black and to ensure that adjacent sides of the tablets in each image did not touch in terms of pixels. This preparation was essential for facilitating subsequent analysis.} This technique delineates the tablet’s outline more prominently against its background, enhancing the contrast without compromising the shape’s integrity. The resulting binary masks effectively highlight the tablet’s silhouette by stripping away any textual or decorative subtleties that could obscure the shape analysis. These masks, depicted in Figure~\ref{fig:mask_preprocess}, offer a simplified yet accurate representation of the tablets’ shapes. The preprocessing of the binary masks is executed in real time, with the output not being stored separately. This approach ensures that the binary mask generation process is integrated seamlessly into the workflow, allowing for immediate analysis without the need for additional storage or handling of preprocessed images. We opted to use binary masks of the entire tablet, which is necessary given the complex and three-dimensional nature of the writing surfaces of cuneiform tablets. While other studies have also utilized shape attributes, such as those on ceramic vessels, they only utilized the vessel profiles, as is customary in ceramic topology. In contrast, our approach requires the full silhouette to account for the intricacies of cuneiform tablets \cite{parisotto2022unsupervised, CARDARELLI2022105640}.

\subsection{Exploratory Data Analysis - Height-to-Width Ratios}
\begin{figure}[htbp]
    \begin{subfigure}[b]{0.46\linewidth}
        \centering
        \includegraphics[width=\textwidth]{images/P469221_masked_image.pdf}
        \caption{Masked image after preprocessing}
        \label{fig:transformation_b_1}
    \end{subfigure}
    \begin{subfigure}[b]{0.46\linewidth}
        \centering
        \includegraphics[width=
\textwidth]{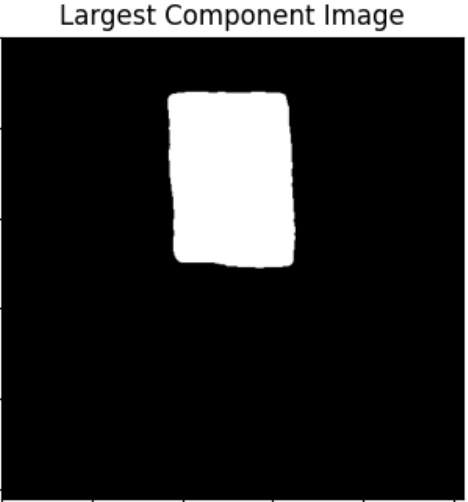}
        \caption{Largest component of the masked image}
        \label{fig:transformation_c_1}
    \end{subfigure}
    \caption{Illustration of the process for extracting the largest component of a tablet: (a) Masked image after preprocessing, and (b) Largest component of the masked image.}
    \label{fig:large_comp_process}
\end{figure}

Our initial investigation aimed to explore potential relationships between the physical dimensions of tablets—specifically their width and height, excluding depth—and their historical periods, hypothesizing that these dimensions could offer insights into the tablets' chronological origins. Given the absence of actual dimensions for most tablets in the dataset, we extracted the largest connected component from the binary masks representing each tablet and derived the pixel height and width of the largest component, a method illustrated in Figure \ref{fig:large_comp_process}. This approach effectively isolates the primary structure of each tablet for more accurate dimension measurement. However, this automated extraction process might introduce inaccuracies due to factors like the merging of other tablet views laid down on the same image or the impact of shadows in the images. Moreover, our dataset includes fragments from certain periods, such as the Neo-Assyrian, which could influence the analysis results. Our technique focuses primarily on the height-to-width ratio, as the measurements we obtained are relative to the tablet's proximity to the camera and the proportions within the original image. For the analysis, we used the Pearson Correlation Matrix, as well as different visualizations~\ref{appendix:class_metrics_desc}.

\subsection{Tablet Period Classification}

Our approach begins with applying traditional machine learning models to assess the predictive utility of simple geometric ratios. Next, we utilize more modern analyses using CNNs to harness image-based features, aiming to improve our classification accuracy by leveraging deep learning algorithms. Our methodology progresses methodically, allowing for a comprehensive evaluation of physical attributes in dating the tablets, as well as the extraction of patterns in shape stylistic features across time and space.

\subsubsection{Evaluation Metrics for Image-Based Period Classification}

In all our experiments, we worked with a dataset containing 94,936 samples. We divided these samples into different segments: Since the models we employ require a considerable amount of data for proper training, 80\% were used for training, 10\% formed the validation set, and the remaining 10\% were allocated for the test set. This division into these proportions was decided upon following previous work ~\cite{9257733, mahmood2023classifying, parisotto2022unsupervised, CARDARELLI2022105640} who used 80\%-90\% for the train set, and 10\%-20\% for the test set, and research regarding dataset splits~\cite{muraina2022ideal}. Due to computational resources, we decided to use a single validation set to help us effectively assess the model's performance, refine hyperparameters, and minimize the chance of overfitting the training data. Furthermore, the validation set allows us to adjust our experimental setup without relying on the test data, maintaining the integrity and objectivity of our final evaluations. For this purpose, we used 10\% of the data, leaving a test set of 10\%.

To evaluate the performance of our classification models in discerning between historical periods of ancient tablets, we utilized the following metrics: Macro F1-score~\cite{grandini2020metrics} that considers our highly unbalanced dataset, Precision and Recall Scores~\cite{davis2006relationship}, Area Under Receiver Operating Curve Score (AUC)~\cite{davis2006relationship} and the Confusion Matrices. 

\subsubsection{Classic Machine Learning models}
Our initial attempt was to assess whether the height-to-width ratio (that was extracted from the masked tablets in the EDA phase) alone could serve as a predictive feature for classifying ancient tablets into their respective historical periods. To establish a baseline for this classification, we opted to create a basic classification model focused exclusively on this single feature. We utilized a Decision Tree model~\cite{quinlan1986induction}. This decision was based on the fact that we are dealing with a single predictor - the height-to-width ratio, and due to the model's ability to handle non-linear relationships that might arise from the dataset's characteristics.

\subsubsection{Unsupervised Feature Extraction} 
For our initial exploration into more advanced models, we utilized DINOv2, a cutting-edge, pre-trained visual transformer model noted for its capabilities in its smaller version (vits14). Originating from the model's introduction~\cite{oquab2023dinov2}, DINOv2 was highlighted as an excellent feature extractor due to its proficiency in meta-learning, enabling it to form robust and generalizable representations of visual features. This quality made it potentially suitable for our task of classifying tablet periods and is a relatively cost-effective solution as we used it solely for inference. Leveraging this pre-trained model, we extracted latent features (384 features per sample) for each masked tablet and each greyscale without further fine-tuning. The extracted features were subsequently input into an Extreme Gradient Boosting (XGBoost) model~\cite{chen2016xgboost} for tablet classification. We determined the XGBoost model's hyperparameters using a 10-fold cross-validation process.

\subsubsection{CNNs}
Building upon our initial classification efforts, we advanced to a more intricate analysis by incorporating image-based features through CNNs. We aimed to verify the potential of physical shapes for determining the chronological origins of the tablets. The architecture of our CNN model consisted of four convolutional layers, each followed by batch normalization layers and pooling, to improve training stability and performance. The network also included two fully connected layers at the end, using a dropout layer, designed to synthesize the features extracted by the convolutional layers into predictions regarding the tablets' historical periods. The hyperparameters chosen for this model are a batch size of 16 samples and a learning rate of $10^{-5}$.\footnote{The hyperparameters were chosen after a process of grid-search, taking into consideration both validation loss and validation accuracy, as well as our computational limitations.} This model configuration was applied to grayscale images and images of the masked tablets. This comparison will quantify the shape's independent contribution to period prediction. We added early-stopping to our training to avoid overfitting. As a result, training for the grayscale images ended after 8 epochs, and for the masked tablet images ended after 9 epochs.

\subsubsection{Advanced CNNs - ResNet50}
In our continued efforts, we implemented a pre-trained ResNe\-t50 model~\cite{he2016deep}, highly regarded in the field for its exceptional performance in image classification challenges \cite{mukti2019transfer, ccinar2021classification, theckedath2020detecting}. This deep residual network is adept at bypassing the vanishing gradient problem, thanks to its 50-layer architecture featuring skip connections. We fine-tuned ResNet50 on two image types: grayscale and masked tablet images (separately).

We fine-tuned the model to predict grayscale and masked images, adjusting the learning rate to $5 \cdot 10^{-5}$ and the batch size to 16.\footnote{The batch size and learning rate hyperparameters were selected through a grid-search method, using a validation set to evaluate each combination of hyperparameters over 3 epochs. We determined the optimal set of hyperparameters based on a balance between computational constraints and performance across evaluation metrics. We used the same structure of the network for both grayscale and masked images to isolate the effect of the change of data.} We used early stopping in the training process to prevent overfitting, with both models' training concluding after 4 epochs.

\subsection{Interpretability and Exploration with VAE}

In the second phase of our research, we shifted our focus towards a deeper understanding of the intricate shape characteristics that distinguish tablets from various historical periods. Our objective was to obtain global-level insights into the defining features of these artifacts, surpassing the limitations of local-focused, model-agnostic interpretability methods. To this end, we opted for explainable generative models, prioritizing a balance between interpretability and ease of training.

After a comprehensive evaluation, we utilized a VAE-based approach. Despite being aware of potential compromises in output fidelity, this decision was driven by the desire to uncover subtle, period-specific shape features, and by previous work~\cite{parisotto2022unsupervised, CARDARELLI2022105640}. VAEs offer a structured, probabilistic method for generating and modeling data. These qualities make them particularly suited for our exploration into the variability of tablet shapes across different historical contexts, and into the latent space of tablet shape characteristics. After consideration and several attempts, we chose to process with the basic VAE and not use any of the methods in the sections above. However, our implementation can still be used together with $beta$-VAE.

\subsubsection{VAE architecture}

In our VAE model designed for the analysis of ancient tablet shapes, the encoder features five convolutional layers with 5*5 size filters that increase in count from 32 to 256, enhancing feature extraction without pooling to maintain spatial detail. At the heart of this model is a 12-dimensional bottleneck, a dense latent space representation of the input images that encapsulates their fundamental features.\footnote{We made our choice of a 12-dimensional latent space by considering both prior research \cite{burgess2018understanding} and the interpretability of the data. In VAEs, there is a trade-off between the size of the latent space and its properties. While a larger space can lead to more accurate reconstructions, it can also make interpreting the latent factors more challenging. Given the relative simplicity of our black-and-white data, we started with a 16-dimensional latent space. However, we were able to achieve good reconstruction quality without significant increases in loss functions (details provided later) by reducing the dimensionality to 12.} Mu ($\mu$) and sigma ($\sigma$), calculated within this bottleneck, define the mean and variance of the latent space, enabling effective sampling through the reparameterization trick. This architecture allows for the precise reconstruction and generation of tablet shapes, 

The decoder part of the VAE reverses the encoder's process, using transposed convolutional layers to gradually upscale the compressed latent representation back to the original image dimensions, enabling accurate reconstruction and novel shape generation. We trained this model over 9 epochs with a 0.0001 learning rate and batch size of 8,\footnote{The selection of these hyperparameters resulted from a grid search, where each combination was assessed over 3 epochs using a validation set. We chose the optimal hyperparameter set by finding a balance between performance on various evaluation metrics and computational limitations.} and included an application of class weights, with weights inversely proportional to the number of samples per class. This VAE architecture adeptly reconstructs and explores the morphology of historical tablets, demonstrating its utility for detailed shape analysis.

\subsubsection{Evaluation Metrics and Loss Functions for the VAE model}
For the VAE model, we used the same train and test set as were used for the masked tablet ResNet50 model, since we wanted to be able to compare the performance on the same dataset. Our VAE model was designed with a focus on interpretability. It incorporates three main components in its loss function:

\begin{itemize}

    \item \textbf{Reconstruction Loss:} This component ensures the VAE's ability to faithfully reconstruct the original tablet images, minimizing the differences between the actual tablets and their VAE-generated equivalents. In contrast to the original VAE paper~\cite{kingma2013auto}, which used  Binary Cross Entropy (BCE) for the reconstruction loss, we opted to use the Mean Squared Error (MSE), as it produced superior results in terms of reconstruction~\cite{CARDARELLI2022105640, khare2021analysis}. 
    \item \textbf{Kullback-Leibler Divergence (KL Divergence):} This measures the information loss when approximating the true data distribution with the model's distribution~\cite{kullback1951information}. By minimizing this divergence, we ensure smooth navigation within the VAE's latent space, facilitating effective latent space exploration.
    \item \textbf{Cross-Entropy Loss:} To enhance the model's classification capabilities and address class imbalances, a weighted classification loss was integrated. This component improves the VAE's ability to distinguish between different historical periods accurately~\cite{doersch2016tutorial}.
\end{itemize}
    
To assess the effectiveness of the VAE model in classifying tablets according to their historical periods, we employed standard evaluation metrics such as precision, recall, macro-AUC score, and macro F1-scores. These metrics and confusion matrices provided a detailed view of the model's performance across different classes and highlighted specific areas where misclassification occurred.

\subsubsection{VAE Bottleneck Layer Exploration}
In our study, the VAE model proved instrumental, mainly through its bottleneck layer, which provides a compressed 12-feature vector representation of each tablet. As a generative model, the VAE moves beyond mere classification, enabling the creation of new tablets that capture the collective characteristics of specific periods. This function allows us to examine not just individual tablets but also to identify period-specific and genre-specific morphological features. Our investigation focused on utilizing the VAE's bottleneck layer to clearly understand the defining morphological characteristics across different historical periods, adopting a systematic approach to analyze the rich representations derived from this layer. Our goal was to elucidate the complex patterns and similarities among the tablets, aiming for a deeper comprehension of their historical and cultural contexts. To achieve this goal, we implemented a series of analytical steps as follows:

\begin{itemize}
    \item \textbf{Prediction Using XGBoost}: We employed the XGBoost algorithm to classify the VAE-encoded representations of tablets, predicting their historical periods. The choice of XGBoost was motivated by its ability to handle unbalanced data. We used the classification to prove that the VAE still holds predictive power regarding tablet dating, and thus can be used for their classification interpretability.
    \item \textbf{Extracting the Mean Tablet Representation:} We derived the average latent representations from our VAE model for tablets classified by historical period, genre, and their intersections. This step enabled us to uncover common visual features among the tablets, setting the stage for subsequent visualizations and analyses. By computing these mean representations, we facilitated the generation of "mean tablet" images within the latent space, effectively synthesizing the archetypal tablet image for each categorized group.
    \item \textbf{Performing Various Clustering Techniques:} We employed a range of clustering techniques, including Hierarchical clustering, on the latent representations to identify underlying patterns and relationships. This analysis enabled us to observe how different periods are grouped at various stages of the algorithm, facilitating an examination of visual connections both within the same millennium and across the entire chronological span of our dataset.
    \item \textbf{Creating Widgets to Traverse the Latent Space:} We developed interactive widgets for a dynamic exploration of the VAE's latent space, facilitating an engaging way to visualize morphological diversity across periods and genres and allow exploration into the meaning of each of the entries of the 12-dimensional vector in the VAE bottleneck layer (see Figure~\ref{fig:widget_bottleneck}).
\end{itemize}
\section{Results} \label{results}
The resulting models and tools visualizing the different tablet features extracted from them, act as a series of strong microscopes and telescopes to measure hypotheses, intended to support human expert evaluation. They aim to propose classifications for cuneiform tablets by period and diplomatic styles and provide aggregate insights on shape-related features across different periods and genres. We believe that these are an initial step into more in-depth localized case studies, within specific periods, archives, and text types.

\subsection{Findings of Exploratory Data Analysis}
\begin{figure*}
\centering
\includegraphics[width=\linewidth]{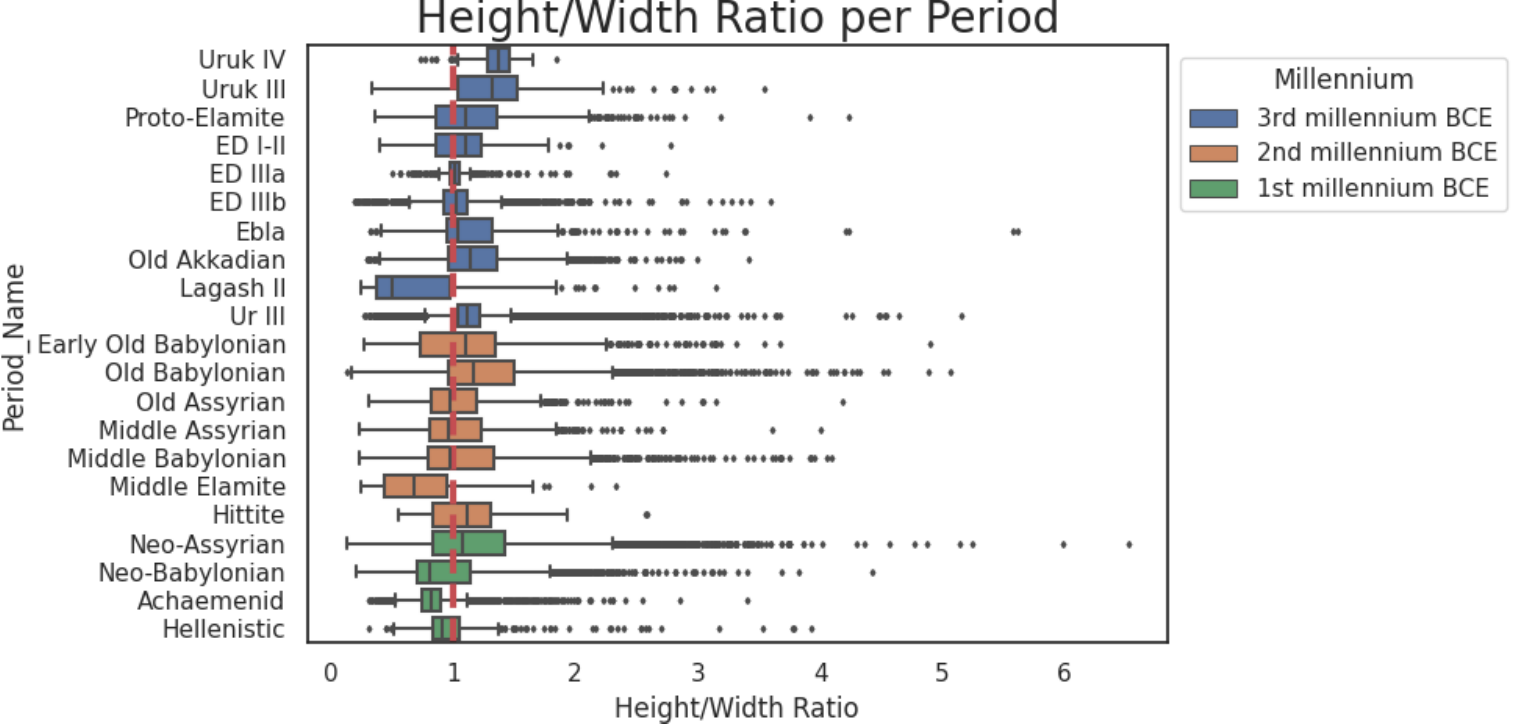}
\caption{The height-to-width ratio across periods is segmented by the millennium. The dashed red line indicates a ratio of height to width equal to 1, distinguishing between portrait and landscape orientations.}
\label{fig:hw_ratio}
\end{figure*}

\begin{figure*}
    \centering
    \begin{subfigure}[T]{0.52\textwidth}
        \centering
  \includegraphics[width=1.0\textwidth]{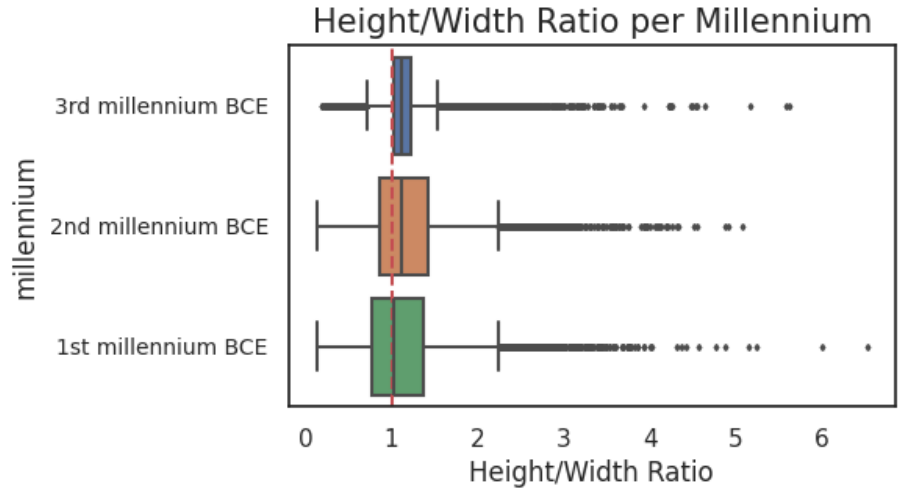}
        \caption{Boxplot of the height-to-width ratio per millennium. The dashed red line indicates a ratio of height to width equal to 1, distinguishing between portrait and landscape orientations.}
        \label{fig:hw_ratio_box}
    \end{subfigure}
    \hfill
    \begin{subfigure}[T]{0.44\textwidth}
        \centering
        \includegraphics[width=0.9\textwidth]{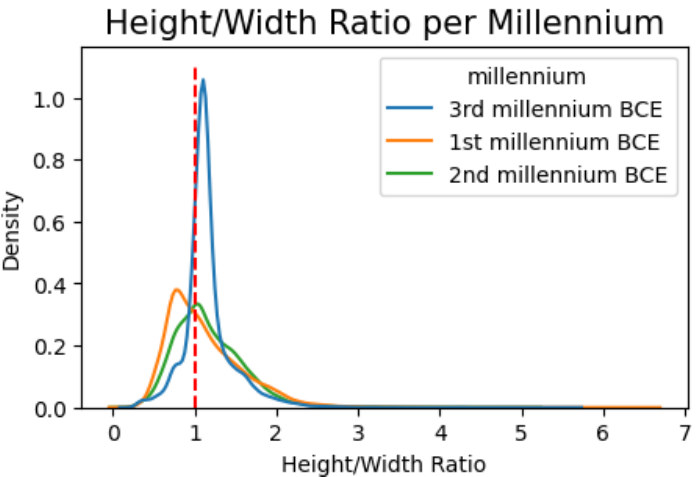}
        \caption{KDE plot of the height-to-width ratio per millennium.}
        \label{fig:hw_ratio_kde}
    \end{subfigure}
    \caption{Analysis of height-to-width ratio per millennium.}
    \label{fig:hw_ratio_analysis_per_mill}
\end{figure*}

Our exploratory data analysis reveals distinct design preferences in cuneiform tablets across different historical periods, as seen in Figure \ref{fig:hw_ratio}. In the Ur III period, with its well known strict bureaucracy, we observed a consistent height-width ratio, showing a tendency for portrait shaped tablets. In contrast, Lagash II, Middle Elamite, and the latter 1st millennium clusters (Neo-Babylonian, Achaemenid, and Hellenistic) displayed a preference for landscape tablets. While on the whole this could be explained by the predominance of certain legal, economic, and administrative genres in later periods, there is a clear statistical tendency to use more landscape format tablets already from the Early Old Babylonian period onwards; with the Old Babylonian period, namely the first dynasty of Babylon, being a clear outlier group.

\begin{figure}[htpb]
\centering
\includegraphics[width=\linewidth]{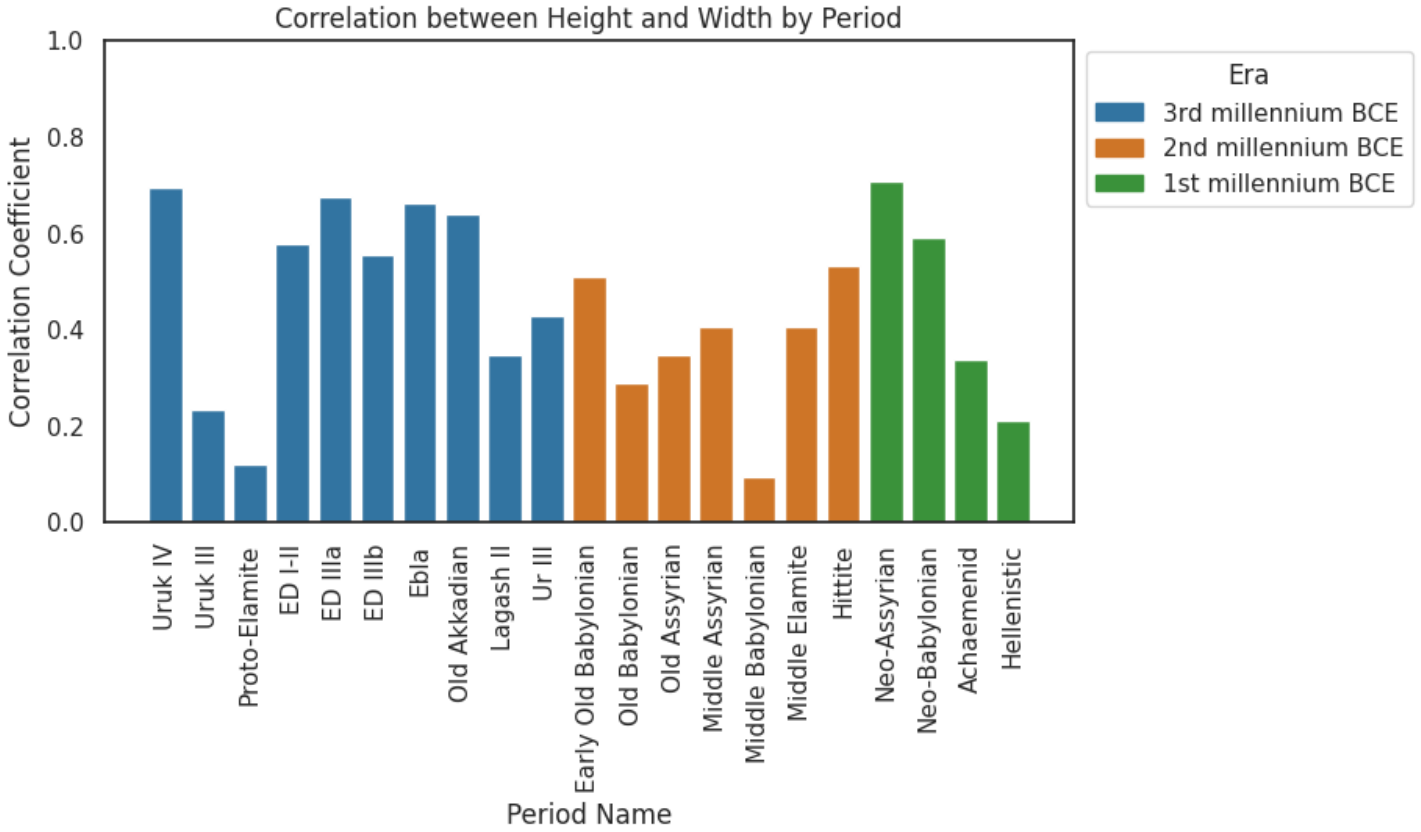}
\caption{Pearson Correlation of height vs. width per period.}
\label{fig:h_w_corr}
\end{figure}

Grouping the tablets by era reveals that those tablets from the 3rd millennium BCE exhibit more consistency in their height-width ratio, often being portrait-shaped, as shown in Figure \ref{fig:hw_ratio_analysis_per_mill}.\footnote{We used a Kernel Density Estimation plot (KDE plot), that visualizes the distribution of data over a continuous interval, offering a smoothed representation of the dataset's density.}
Analyzing the Pearson correlation~\ref{fig:h_w_corr} between height and width revealed a positive linear relationship in certain periods, with some showing correlations above 60\%. However, other periods exhibited weak correlations, indicating a more complex relationship between tablet dimensions and their historical context. While the majority of genres showed a correlation exceeding 40\%, the significant imbalance in the dataset across different genres limited further genre-based analysis.

\subsection{Tablet Period Classification Models}

The application of the Decision Tree model, utilizing only the height-to-width ratio of the tablets, yielded a macro-F1 score of 8\%. This extremely low score highlights the challenges in using the height-to-width ratio as the sole feature for accurately classifying the tablets into their historical periods and underscores the necessity for integrating more sophisticated and comprehensive shape-related features into the classification model. Interestingly, the model was not able to learn at all when it came to the grayscale tablets, scoring a macro-F1 score of only 2\%.

Using the features extracted from DINOv2, the XGBoost model achieved a macro F1-score of 41\% for the masked tablet images, marking a significant improvement over the previous Decision Tree model. Considering that DINOv2 had never been exposed to our dataset before, this performance was quite remarkable. We then proceeded with additional classification models that had been directly trained on our specific dataset.

The deployment of our CNN model on the classification task yielded a Macro F1-score of 61\% for grayscale images and 49\% for silhouette images of the tablets. These results substantially surpass the performance of initial models that used only the height-to-width ratio for classification.
Significantly, certain periods showed exceptional model performance given only the masked tablet images as input; the Ur III period reached a Macro F1-score of 84\%, and the Neo-Assyrian period had a remarkable 92\% Macro F1-score.

\begin{figure*}
\centering
\begin{subfigure}{.5\textwidth}
\centering
\includegraphics[width=\linewidth]{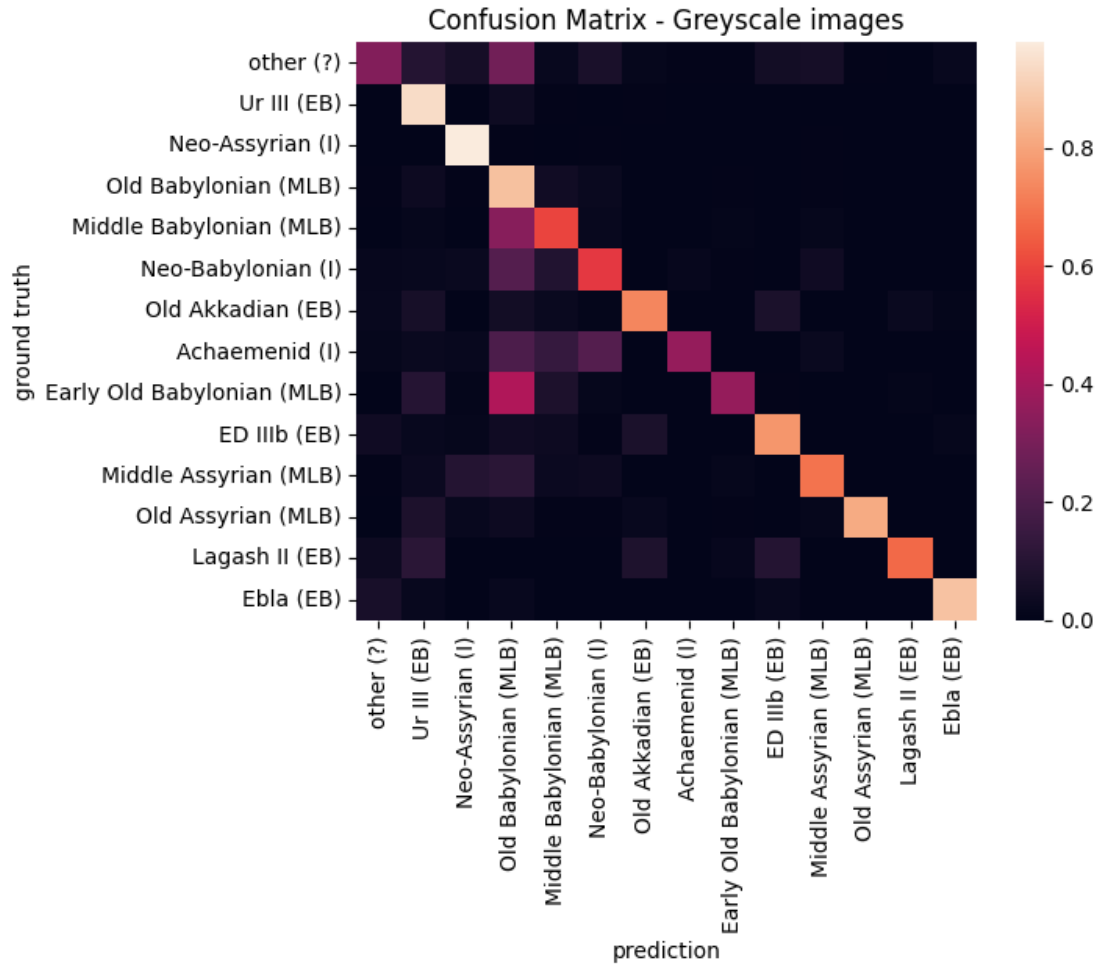}
\caption{Grayscale Images}
\label{fig:greyscale_conf_mat}
\end{subfigure}%
\begin{subfigure}{.5\textwidth}
\centering
\includegraphics[width=\linewidth]{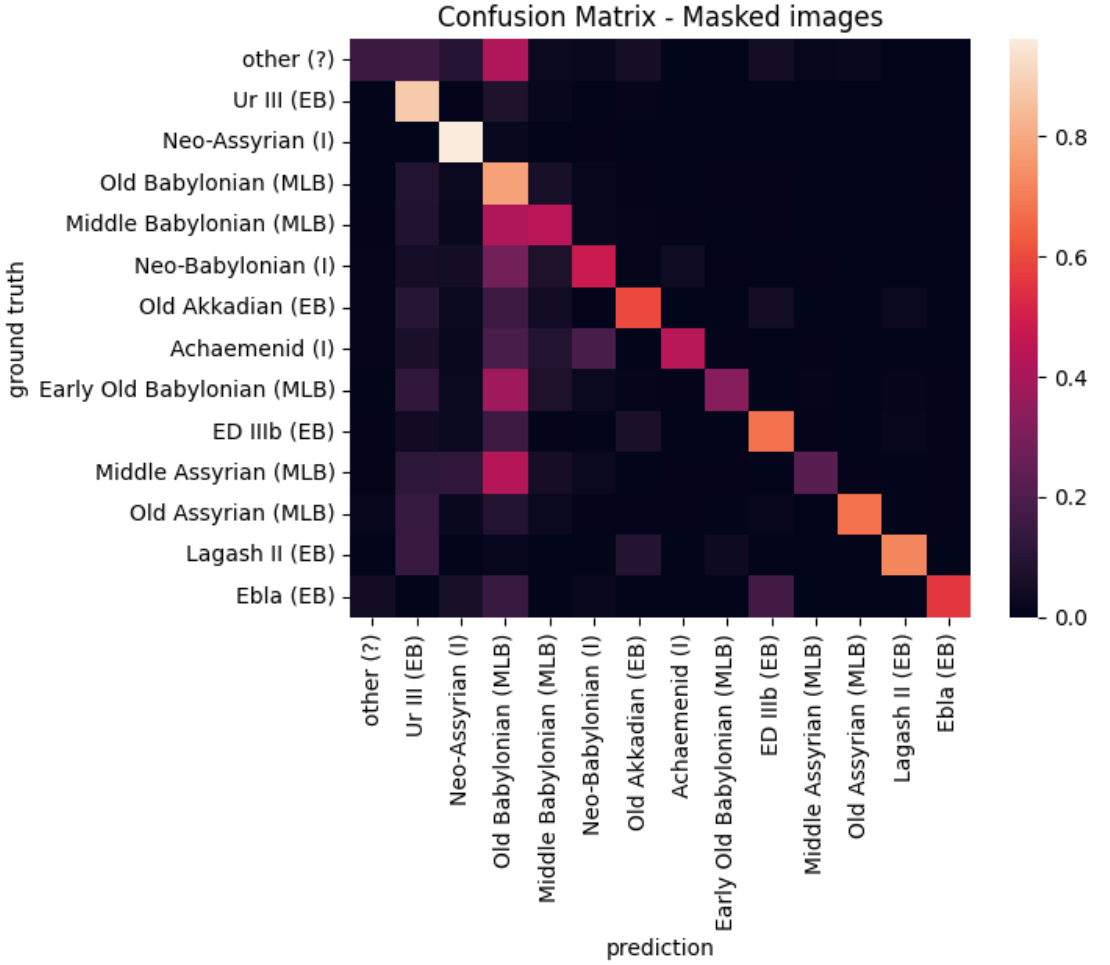}
\caption{Masked Images}
\label{fig:masked_conf_mat}
\end{subfigure}
\caption{Confusion matrices comparing the classification performance of ResNet50 on different types of images}
\label{fig:confusion_matrices}
\end{figure*}

When fine-tuned on grayscale images, ResNet50 achieved a macro average F1-score of 71\%. The confusion matrix, shown in Figure \ref{fig:greyscale_conf_mat}, reveals slight confusion between closely related historical periods, but overall a low level of misclassification. 

The performance on masked images remained solid, with the model achieving a 61\% Macro F1-score. The confusion matrix for the masked images, depicted in Figure \ref{fig:masked_conf_mat}, indicates a challenge in differentiating certain periods, such as the Old Babylonian. It is worth noting that the ResNet50 model showed remarkable accuracy in distinguishing specific periods from masked tablet images only, achieving F1 scores 77\% F1 score for the Old Assyrian period, 87\% F1 score for the Ur III period, and 95\% F1 score for the Neo-Assyrian period.

Table \ref{tab:class_res_sum} compiles the results from the different models used, with a note that classes with fewer than ten instances in the test set were aggregated into an "Other (?)" category for clarity.

\begin{table*}[ht]
\centering
\caption{Tablet Dating - Results Summary}

\sisetup{detect-all}
\NewDocumentCommand{\B}{}{\fontseries{b}\selectfont}

\begin{tabular}{
  @{}
  l
  S[table-format=1.2]
  S[table-format=1.2]
  S[table-format=1.2]
  S[table-format=1.2]
  S[table-format=-1.2]
  S[table-format=1.2]
  S[table-format=-1.2]
  S[table-format=1.2]
  @{}
}
\toprule
Models & \multicolumn{8}{c}{Metrics} \\ 
\cmidrule(l){2-9}
{}&\multicolumn{2}{c}{Macro-Precision} & \multicolumn{2}{c}{Macro-Recall} & \multicolumn{2}{c}{Macro-OvR AUC} & \multicolumn{2}{c}{Macro-F1} \\
\cmidrule(lr){2-3} \cmidrule(lr){4-5} \cmidrule(lr){6-7} \cmidrule(l){8-9}
& {Grayscale} & {Masked} & {Grayscale} & {Masked} & {Grayscale} & {Masked} & {Grayscale} & {Masked} \\
\midrule
Decision Tree &  {-} & {0.11} & {-} & {0.15} & {-} & {0.64} & {-} & {0.08} \\
DINOv2-vits14 &  {0.04} & {0.63} & {0.06} & {0.36} & {  0.54} & {0.88} & {0.02} & {0.42} \\
CNN & 0.74 & 0.61 & 0.54 & 0.44 & 0.93 & 0.88 & 0.61 & 0.49 \\
ResNet50 & {\textbf{0.77}} & {\textbf{0.70}} & {\textbf{0.68}} & {\textbf{0.56}} & {\textbf{0.97}} & {\textbf{0.93}} & {\textbf{0.71}} & {\textbf{0.61}} \\
\bottomrule
\end{tabular}
\label{tab:class_res_sum}
\end{table*}

\subsection{VAE Results}
Following the initial classification results, we sought to interpret the underlying characteristics responsible for the model's predictions. To achieve this, we employed a VAE neural network. 

\begin{figure}[htbp]
\centering 
\includegraphics[width=\linewidth]{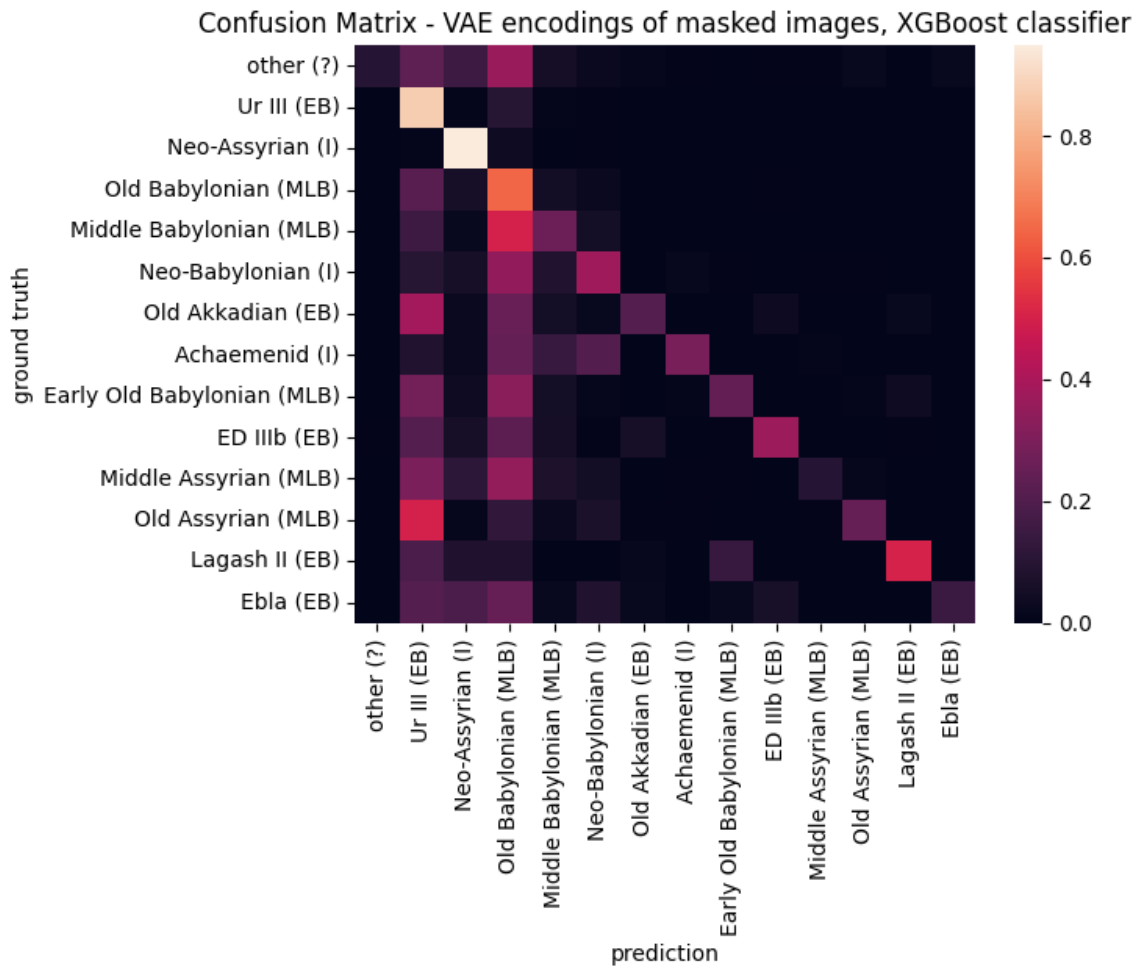}
\caption{Confusion Matrix - VAE encodings classification using XGBoost.}
\label{fig:masked_xgb_conf}
\end{figure}

\subsubsection{VAE Bottleneck Period Classification}
The deployment of our VAE model has provided valuable insights into the connections between tablet shapes and their historical periods, with an overall macro F1-score of 43\%, and a macro AUC score of 88\%. Classification outcomes were achieved via an XGBoost model, as detailed in Appendix~\ref{appendix:class_metrics_desc}, showcasing the model's adeptness at conserving critical classification information within a 12-dimensional bottleneck representation. Impressively, the model scored Macro F1-scores of 79\% for the Ur III category and 92\% for the Neo-Assyrian category, indicating its predictive capability for these periods. Further examination of the confusion matrix (Figure \ref{fig:masked_xgb_conf}) uncovers significant confusion mainly between the Ur III, Old Babylonian, and the rest of the classes.

\subsubsection{Analysis of the VAE Bottleneck}
 By examining the mean encoding (12-features retrieved from the bottleneck layer of the VAE) for each period\footnote{We obtained the average tablet representation for each group by retrieving the bottleneck layer from the VAE, averaging the 12-entry vectors, and passing the mean vector through the VAE decoder.} (Figure \ref{fig:mean_tab_by_per}) we can potentially uncover shared visual traits and features that differentiate tablets across historical eras. This approach unlocks a deeper understanding of tablet shapes and their connection to specific periods. 

Additionally, analyzing genre-specific average tablets \\(Figure \ref{fig:mean_tab_by_gen}) allows us to expose shape trends within genres, opening new avenues for understanding historical trends and potentially distinguishing tablets based on both period and genre.

Following the generation of mean tablets for each period and genre, we utilized hierarchical clustering to explore their relationships. Specifically, we analyzed each distinct genre group individually, using their mean period vector representations as the clustering input.

\begin{figure*}[ht]
\centering
\includegraphics[width=\linewidth]{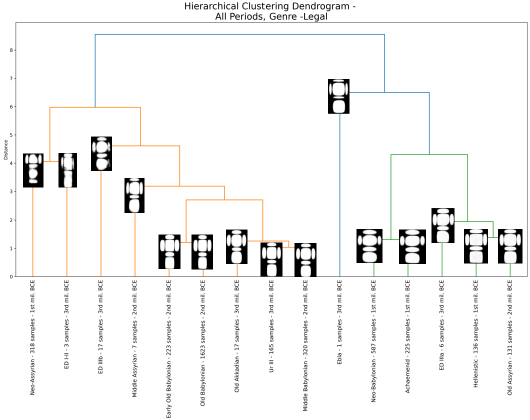}
\caption{Dendrogram of Mean Period Tablet - Legal Genre. The colors represent the clusters determined by the Hierarchical clustering.}
\label{fig:legal_dendro}
\end{figure*}

\begin{figure*}[ht]
\centering
\includegraphics[width=\linewidth]{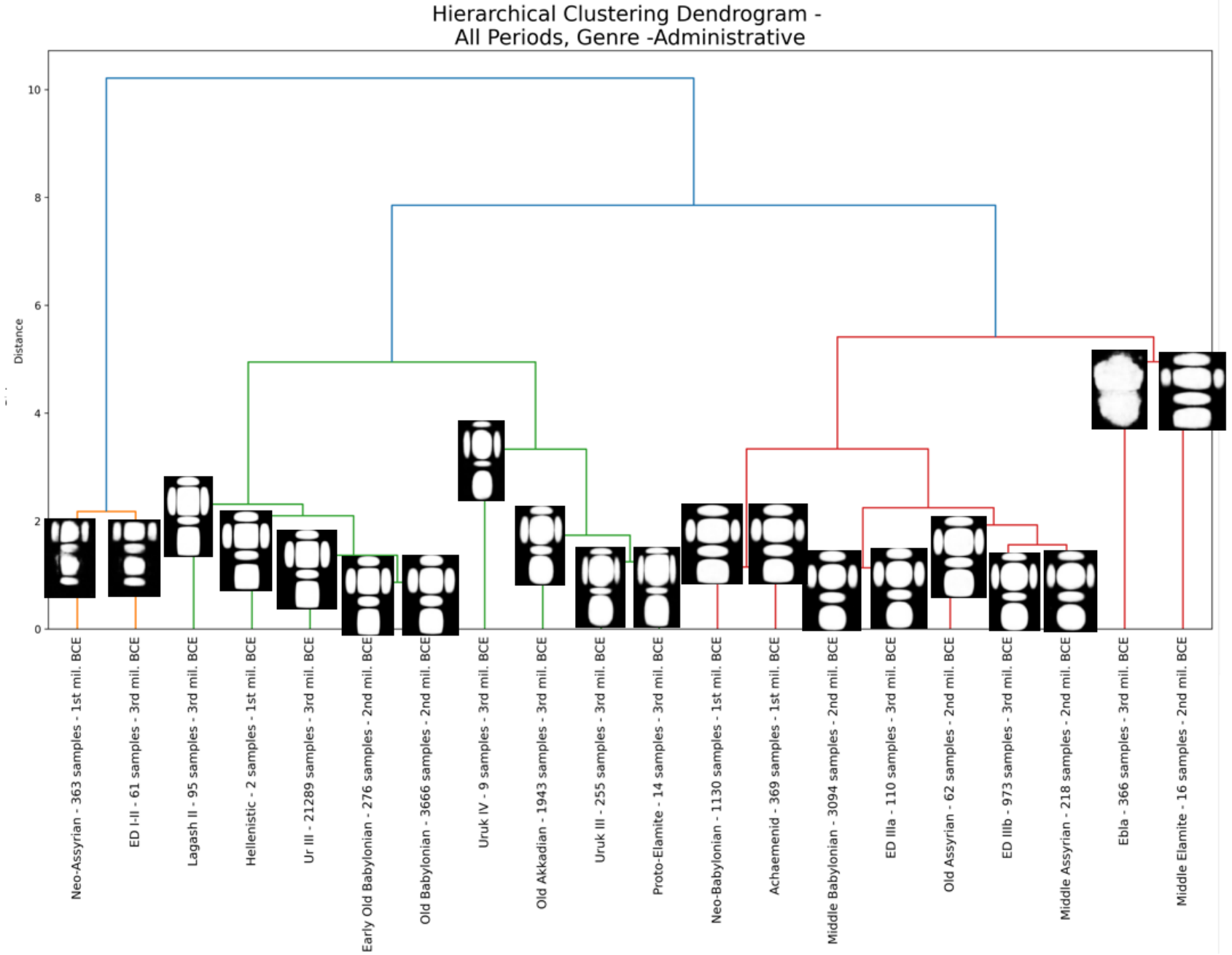}
\caption{Dendrogram of Mean Period Tablet - Administrative Genre. The colors represent the clusters determined by the Hierarchical clustering.}
\label{fig:admin_dendro}
\end{figure*}

Hierarchical clustering of tablets from the Legal genre (Figure \ref{fig:legal_dendro}) shows chronological clustering, where periods close in time, like the Early Old Babylonian and Old Babylonian periods, cluster together. This pattern is consistent, with a notable cluster combining periods mostly from the 2nd millennium BCE, and a similar pattern observed in the 1st millennium BCE with the Neo-Babylonian and Achaemenid periods. In the Administrative genre (Figure \ref{fig:admin_dendro}), tablets with round shapes cluster together, irrespective of their era. A distinct cluster includes tablets photographed differently, where the top view is positioned at the bottom of the image.

We continued our analysis by producing separate dendrograms per millennium, to get a closer look at the similarities in each era, as can be seen in Figure \ref{fig:dendro_by_mill_admin} which shows the Administrative genre dendrograms, divided by millennium. We can see nicely, how similar-shaped average-tablets are clustered together, and how close chronological periods are clustered together like the Middle Babylonian and the Middle Assyrian periods, and Old Babylonian and Early Old Babylonian which were clustered together.

\begin{figure}[htbp]
\centering 
\includegraphics[width=\linewidth]{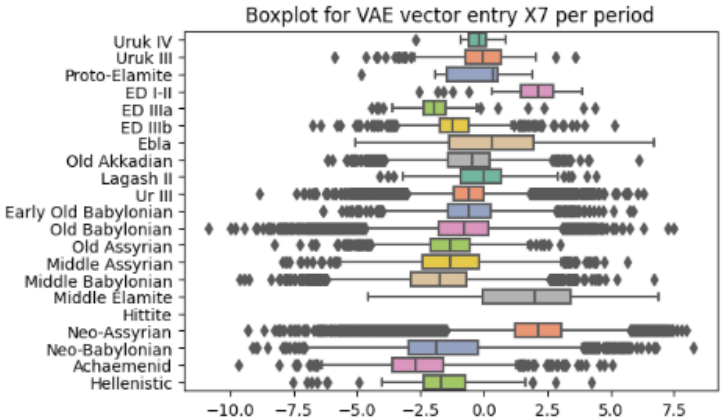}
\caption{Summary of VAE bottleneck features: Focus on the 12th entry of the bottleneck vector across different periods.}
\label{fig:entry_7}
\end{figure}

We conducted a deeper analysis of the mean tablets by exploring the individual components within their 12 feature vector representations retrieved from the VAE. Initially, we plotted the values of various entries for each period, as illustrated in Figure \ref{fig:entry_7}. This plot reveals that the Middle Elamite, ED I-II and Neo-Assyrian periods exhibit higher values for this particular entry, whereas the Achaemenid period, tends to maintain lower values for the same entry. This entry seems to be in charge of the way the different angles of the tablet were placed - is the top view of the tablet presented on the top or the bottom of the image. Entry no. 8 of the latent vector seems to be in charge of how thick is the tablet from a side view, and entry no. 9 seems to be in charge of how thick the tablet is from the top/bottom. 

As we wish to provide tools for researchers to further dive into the meaning of the different VAE components, thus understanding the important factors in the cuneiform tablet shape-based tablet dating, we provided two visual widgets. 

The first widget showcases a random tablet from a selected period and genre, treating each component of the tablet's VAE bottleneck representation as a control knob. As demonstrated in Figure \ref{fig:widget_bottleneck}, this enables the user to adjust the value of a specific element while keeping the rest unchanged.\footnote{Once a user changes a value for one of the features, a new vector is created, and passed through the VAE decoder, which generates an appropriate tablet.} Although each element influences multiple visual features, certain latent features play more distinctive roles and can be better explained.

The second widget, illustrated in Figure \ref{fig:interpulation_widget}, permits users to navigate the latent space generated by the VAE between two mean tablets from chosen periods. Users can modify the interpolation factor to select their position along the linear combination, offering insights into the evolution of tablet shapes over time or facilitating comparisons of tablets from identical chronological periods in different regions.

\section{Discussion} \label{discussion}

Improving the standardization of the process of dating cunei-form tablets enhances our understanding of ancient civilizations, responding to the increasing interest prompted by digitization projects that make these artifacts widely accessible. Our study introduces a novel approach for classifying cuneiform tablets by historical periods through shape analysis, employing advanced machine-learning techniques to aid archaeologists and historians. This method is designed to complement expert analysis by uncovering insights into the design and historical context of the tablets, particularly how their shapes reflect their functions and the societies that created them. To achieve our goals, our work was structured in several stages:

First, our exploration of the dataset revealed its diversity, with most tablets adopting a "fat-cross" shape. However, we observed deviations from this general pattern, particularly in specific periods and genres. Notably, during the Lagash II period, artifacts within the Royal/Monumental genre primarily consisted of cylindrical shapes, unlike the common tablet forms. Similarly, the Neo-Assyrian period was characterized by numerous small tablet fragments. These variations pose challenges for tracking the evolution of tablet designs and comprehending their historical and cultural significance.

Additionally, we identified discrepancies in how some tablets were represented, primarily due to the positioning of tablet angles in images on a single canvas during dataset curation. We also learned that our dataset is unbalanced, with over 75\% of the samples belonging to four classes and over 45\% of the tablets belonging to the Administrative genre (as can be seen in Figures~\ref{fig:nsamples_per_period} and \ref{fig:nsamples_per_genre}). This observation underscored the necessity for a standardized approach to collecting tablet images, as well as the importance of excavating and digitizing as many tablets from as many periods as possible. Standardization and variability would enable more accurate analysis and minimize potential biases impacting automated shape analysis methods.

Second, we focused on the height-to-width ratio of the tablets to get initial conclusions regarding the shape contribution to the tablet period. We realized that within specific periods, this ratio shows remarkable consistency, suggesting standardized production methods reflective of the period's administrative and cultural practices. For instance, during the Uruk IV, ED IIIa, Ur III, and Achaemenid periods, we observe a high degree of uniformity in the height-width ratios of tablets. On a broader scale, our dataset suggests a general trend toward a consistent overall shape across all periods, demonstrating a commonality in design preferences or functionalities throughout history, normally choosing quadrilateral-shaped tablets. However, we could not detect a consistent overall trend across all periods, which underlines the complexities of tablet design evolution, hinting at the interplay of various factors— technological advancements, changes in administrative needs, or shifts in cultural values—that influenced tablet shape and size.

Using solely the height-to-width ratio in a Decision Tree model to classify cuneiform tablets by historical period has shown that while these basic measures can provide some insights, they are not sufficient for detailed analysis, as using them in a Decision Tree only scored 8\% macro F1-score. Incorporating more advanced shape features into our models significantly improved our ability to distinguish between periods, offering a more accurate and intricate understanding of the tablets' design and historical context.

Third, we employed advanced machine learning and deep learning techniques to evaluate the effectiveness of two-dimen-sional representations in classifying cuneiform tablets into periods. Our focus was particularly on the extent to which silhouette representations, created using black and white masks, could encapsulate essential information for this classification task. Among the models we tested, ResNet50 demonstrated superior performance, showing strong accuracy for the most frequently occurring classes and providing satisfactory results overall. This suggests that while shape is a significant factor in dating tablets, it is not the sole determinant. Through the analysis of confusion matrices, we observed that some tablets from the same millennium or geographical area often were misclassified as one another, indicating the persistence of design traditions within these groups. In contrast, periods characterized by unique and innovative design features were more distinctly identifiable. 

Fourth, we used VAEs to extract 12 features from each tablet's silhouette image through the bottleneck layer. This method allowed us to classify tablets into their historical periods using just these 12 features and to gain new insights into the distinct characteristics of tablets from various historical periods. When we applied an XGBoost model to these VAE-generated features, we expected a lower performance compared to ResNet50 because VAEs focus on different loss functions and are traditionally unsupervised models. Nonetheless, for major classes in our dataset, such as the Neo-Assyrian and Ur III periods, the model achieved impressive scores. The results underscore the VAE's potential as a tool for data representation in shape-related classification tasks, especially in cuneiform tablet studies. 

\begin{figure}[htbp]
\centering 
\includegraphics[width=\linewidth]{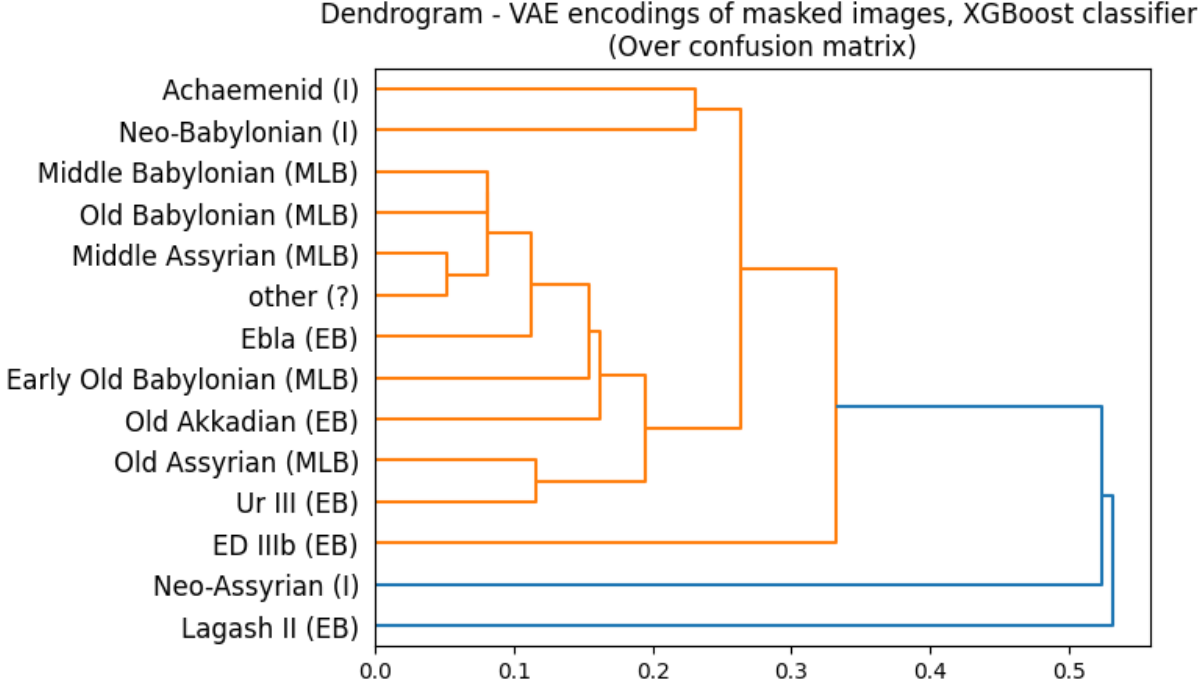}
\caption{Dendrogram - VAE encodings of masked images, (Over confusion matrix), showing which classes were more often confused during training.}
\label{fig:heir_vae_xgb}
\end{figure}

Fifth, we utilized the latent vectors generated by the VAE for aggregative visualizations and analyses across different historical periods. An interesting finding was observed in the hierarchical clustering dendrogram that was created based on the confusion matrix from the XGBoost model trained on these VAE latent vectors (Figure \ref{fig:heir_vae_xgb}). This analysis showed a close classification among classes related to Babylonia, indicating similarities in their features. In contrast, the Neo-Assyrian and Lagash II periods were distinctly separated, likely due to their unique shapes compared to the rest of the dataset. 

Sixth, we calculated the average tablet shape for each period and genre, as well as for combinations thereof, and conducted various analyses. These analyses demonstrated that genres such as Administrative and Legal tended to maintain consistent shapes across different periods, as shown in Figure \ref{fig:mean_tab_by_gen}, which suggests a correlation between their functions and visual identities. On the other hand, genres like Royal/Monumental, Private/Votive, and letters displayed more variation in their average representations, indicating a more specific relationship with shape over time. Similarly, when calculating the mean tablet per period, the Ebla, Neo-Assyrian, and Middle Elamite periods exhibited high shape variability, as depicted in Figure~\ref{fig:mean_tab_by_per}. 

Seventh, we also performed hierarchical clustering of the average shapes of tablets, filtering the analysis by genre. The clustering results within the Legal genre indicated a continuity in legal practices over closely related periods, particularly evident from the 2nd to the 1st millennium BCE. The dendrogram for the Administrative genre reveals that the formation of certain clusters is influenced by the representation of the tablet images, notably the orientation with the top view of the tablet placed at the bottom. This indicates that the clustering algorithm may be picking up on features related to the image orientation rather than the inherent shape of the tablets themselves. Additionally, the dendrogram shows that tablets from the same millenniums tend to cluster together. This could suggest that there are shared characteristics or styles in tablet design within these time frames, reflecting continuity in administrative practices or record-keeping conventions across those periods. 

Lastly, we developed two interactive widgets. These tools are designed to enable researchers to conduct a more specific investigation into the evolution of tablet shapes between different periods and to explore the meaning of each feature within the 12-feature latent vector derived from the VAE.


\section{Conclusions} \label{conclusions}

Cuneiform tablets, the earliest form of writing, provide insights into ancient socio-economic, legal, and cultural practices. The evolution of their designs reflects technological progress, administrative changes, and shifts in cultural values. Recent digitalization efforts have enabled open access to these tablets, shifting analysis from manual to statistical methods and enabling a broader, more efficient examination.

Our research aims to enhance the accuracy of classifying cuneiform tablets by historical period through shape analysis, employing advanced machine learning techniques to assist archaeologists and historians. This approach does not seek to replace human expertise but to complement it by revealing less apparent insights into tablet designs and their historical contexts.

We focus on using ResNet50 for shape-based dating and VAE for feature extraction and analysis, selected for their ability to recognize subtle characteristics of tablets from different periods. Our dataset includes images of tablets spanning various eras, representing a significant advancement in the field.

By moving towards automated statistical analysis, our study proposes a new methodology for analyzing cuneiform tablets, standardizing artifact dating, and exploring the relationship between tablet shape and format and its historical significance. Despite challenges like dataset imbalances and irregular tablet shapes, our work marks a considerable contribution to historical and archaeological research, enriching the toolkit available for studying ancient civilizations through computational methods.

\section{Code and Data Availability} 
\label{code_data_avail}

The data and code used in this study are available from the authors upon request.

\phantomsection
\section*{Acknowledgments} 

\addcontentsline{toc}{section}{Acknowledgments} 

We want to thank Ben-Gurion University of the Negev and the Department of Software and Information Systems Engineering for the resources that were instrumental in conducting this research. We are also thankful to Morris Alper for his role in initiating this project and for his support in the initial phase.

Additionally, we acknowledge the assistance of generative large language models~\cite{ChatGPT, gemini}, which were used to enhance the clarity and richness of the text in this paper.
\phantomsection
\bibliographystyle{unsrt}
\bibliography{sample}

\appendix
\begin{appendices}

\section*{Appendix}
\subsection*{Metrics Description} \label{appendix:class_metrics_desc}
In this section, we delve deeper into the metrics utilized to evaluate our models:

\begin{itemize}

    \item \textbf{Precision and Recall Scores:} These scores are examined individually for each category to understand the model's performance per period, identifying which categories are more challenging to classify accurately.
    
    \item \textbf{Macro F1-Score:} This metric offers a balanced evaluation of the model's macro-precision (the accuracy of identified classifications) and macro-recall (the model's capacity to identify all pertinent instances) scores, with an average taken across all classes. Importantly, it does not consider the number of samples within each class, which helps us address potential biases toward the larger classes in our unbalanced dataset.
    
    \item \textbf{Area Under the Receiver Operating Characteristic Curve (AUC):} The AUC score is measured as the area under the receiver operating characteristic (ROC) curve. it is utilized to compare different models by quantifying a model's ability to distinguish between categories. We used the score in a one-vs-all manner for multiclass classification. It's worth noting that this measurement is biased towards the larger classes in the dataset.
    
    \item \textbf{Confusion Matrices:} To gain further insights into the model's classification behavior, confusion matrices are used. These matrices reveal patterns in misclassification by showing the confusion between categories and providing a visual representation of classification errors.
\end{itemize}
\newpage
\onecolumn

\subsection*{VAE representations of the mean tablets}
\begin{figure*}[h!]
\centering 
\includegraphics[width=.9\linewidth]{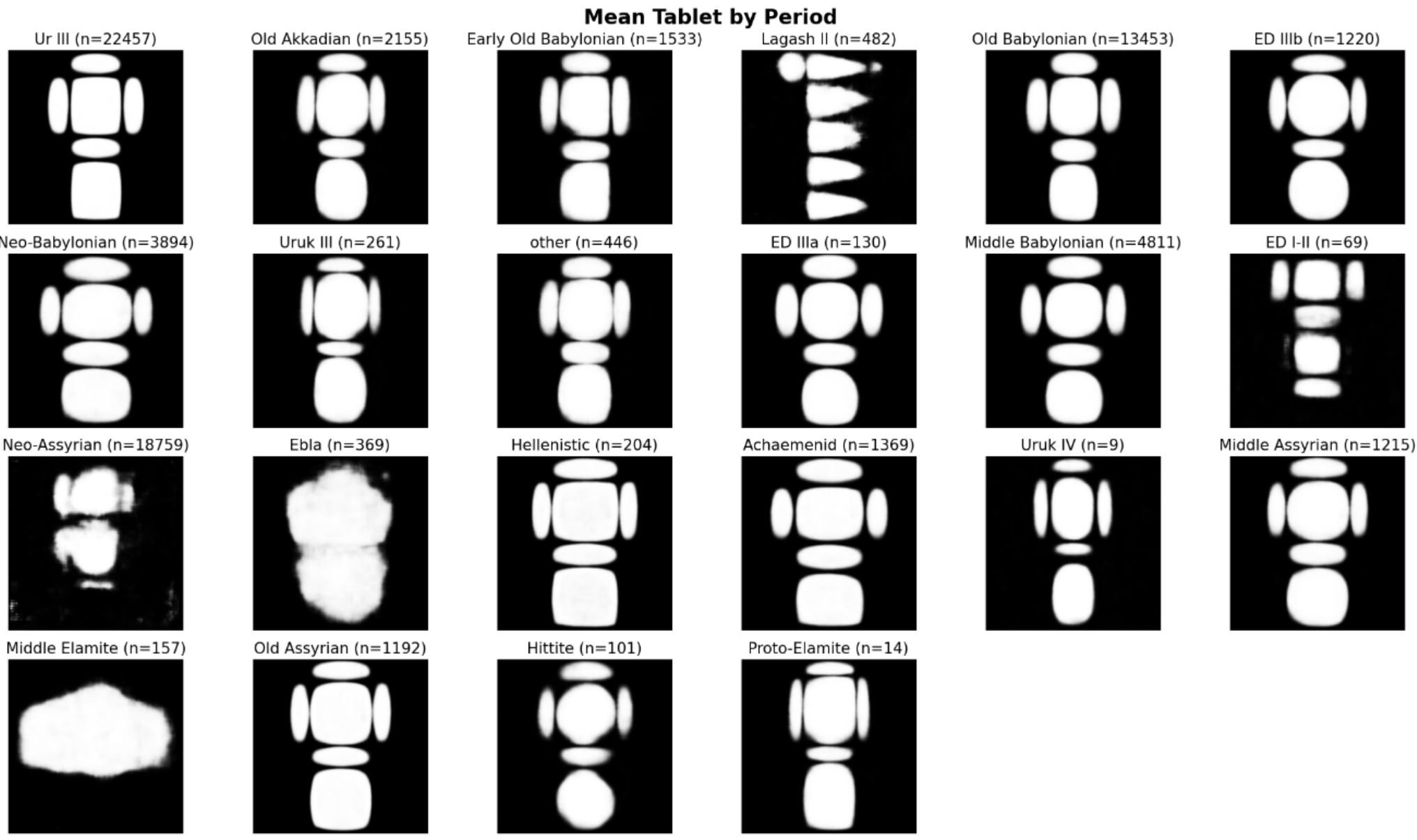}
\caption{Mean tablet by period, including the number of samples per period.}
\label{fig:mean_tab_by_per}
\end{figure*}
\begin{figure*}[h!]
\centering 
\includegraphics[width=.9\linewidth]{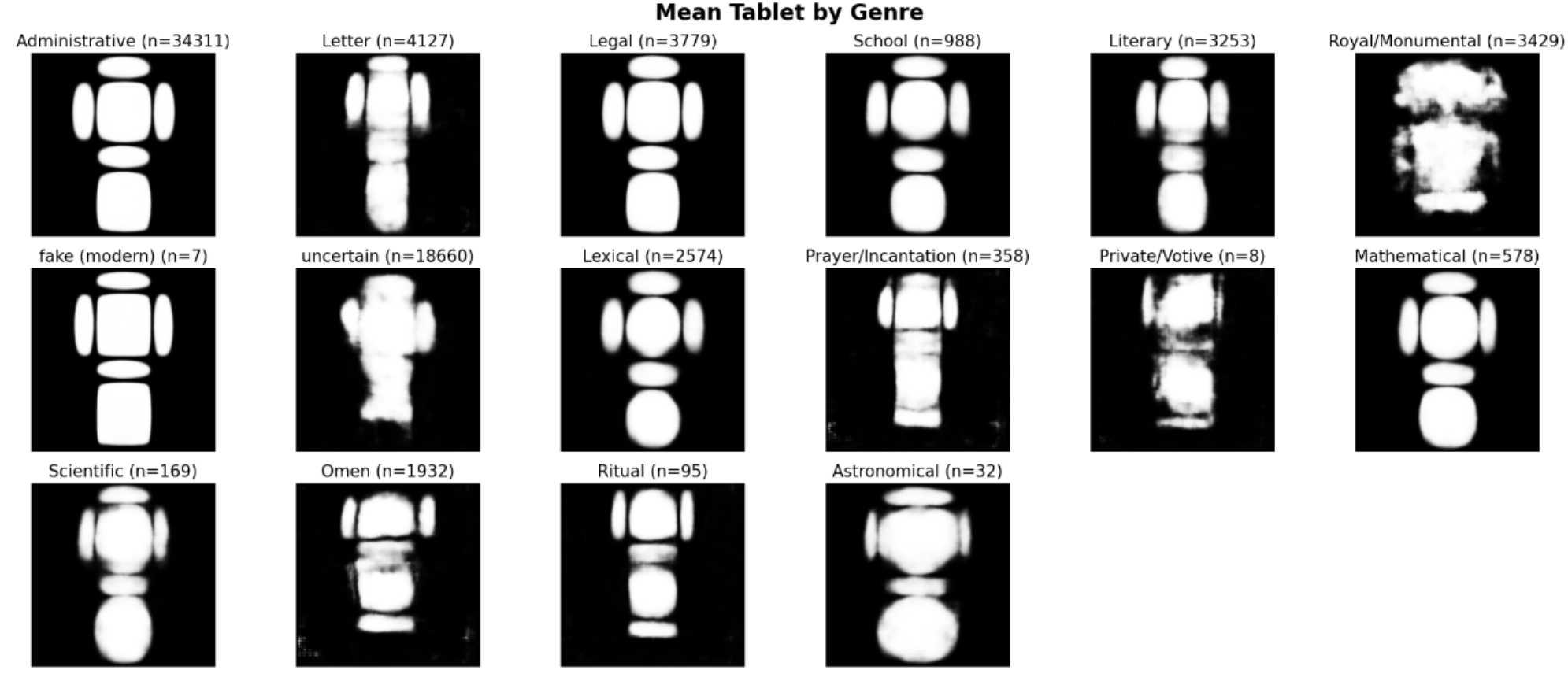}
\caption{Mean tablet by genre, including number of samples per genre.}
\label{fig:mean_tab_by_gen}
\end{figure*}
\newpage
\onecolumn
\subsection*{Dendrograms from Hierarchical Clustering over VAE latent vectors - Administrative genre}

\begin{figure*}[hbt]
    \centering
    \begin{subfigure}[T]{0.31\textwidth}
        \centering
        \includegraphics[width=\textwidth]{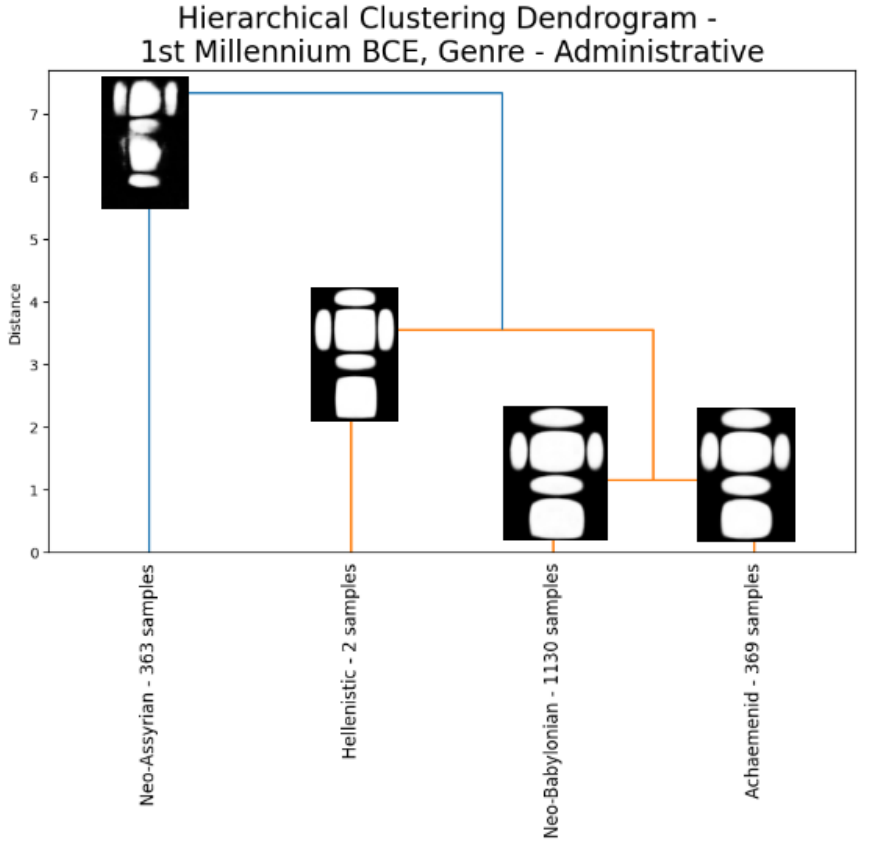}
        \caption{1st Millennium BCE}
        \label{fig:dendro_mill1_admin}
    \end{subfigure}
    \begin{subfigure}[T]{0.31\textwidth}
        \centering
        \includegraphics[width=\textwidth]{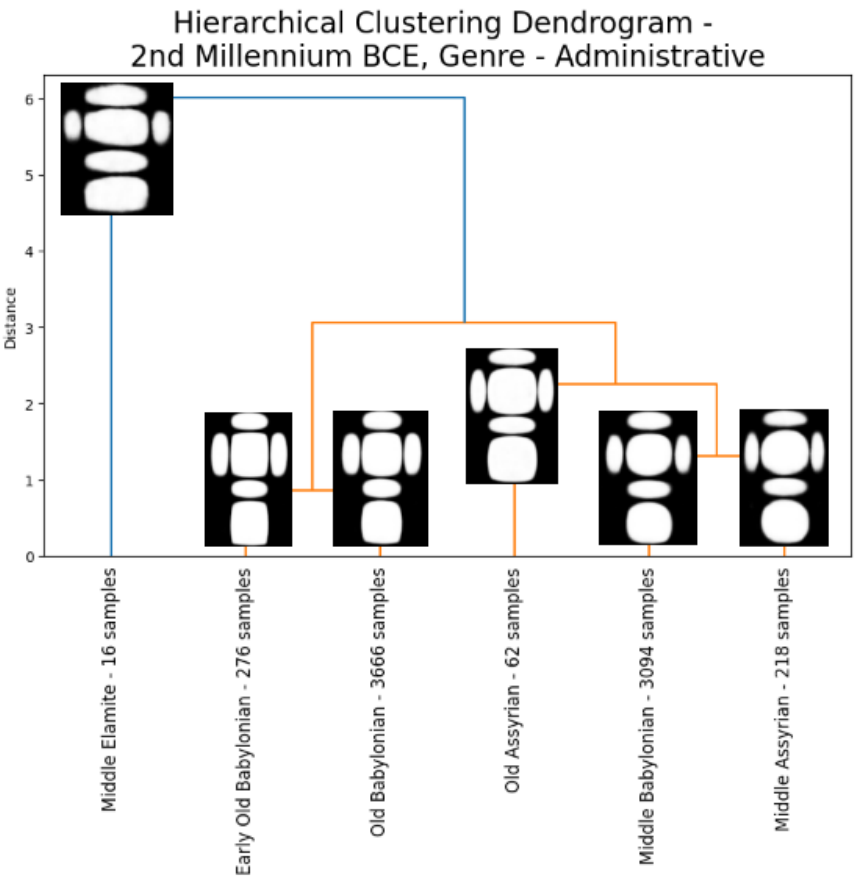}
        \caption{2nd Millennium BCE} 
        \label{fig:dendro_mill2_admin}
    \end{subfigure}
    \begin{subfigure}[T]{0.31\textwidth}
        \centering
        \includegraphics[width=\textwidth]{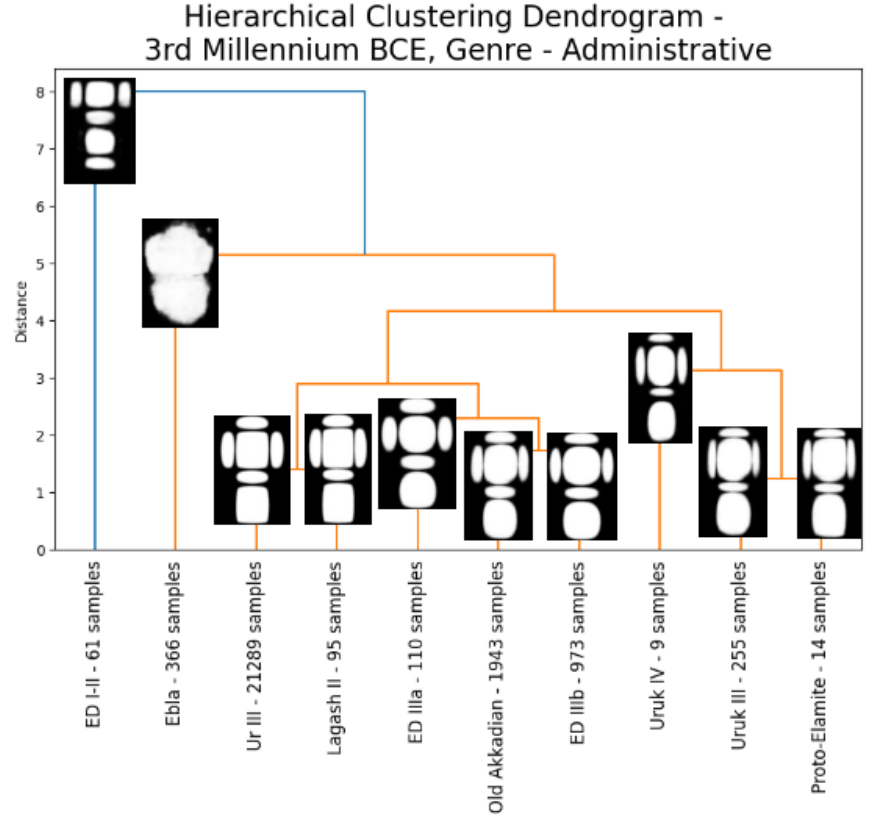}
        \caption{3rd Millennium BCE}
        \label{fig:dendro_mill3_admin}
    \end{subfigure}
    \caption{Dendrograms of hierarchical clustering by millennium for tablets in the Administrative genre. The colors represent the clusters determined by the hierarchical clustering.}
    \label{fig:dendro_by_mill_admin}
\end{figure*}
\end{appendices}

\end{document}